\documentclass[conference]{IEEEtran}

\usepackage[T1]{fontenc}
\usepackage{microtype}
\usepackage{booktabs}
\usepackage{multirow}
\usepackage{graphicx}
\usepackage[normalem]{ulem}
\usepackage{makecell}
\usepackage{amssymb}
\usepackage{amsfonts}
\usepackage{algorithmic}
\usepackage{amsthm}
\usepackage{dsfont}

\newtheorem{theorem}{Theorem}

\usepackage[english]{babel}
\usepackage{subcaption}
\usepackage{soul}
\usepackage{hyperref}
\usepackage[ruled,linesnumbered]{algorithm2e}

\newcommand{\argmax}{\operatornamewithlimits{argmax}}

\newcommand{\name}{\text{TextGuard }}
\newcommand{\namenospace}{\text{TextGuard}}

\newcommand{\update}[1]{\textcolor{black}{#1}}
\newcommand{\revision}[1]{\textcolor{black}{#1}}

\usepackage{tikz}

\usepackage[cmex10]{amsmath}

\begin{document}

\title{\namenospace: Provable Defense against Backdoor Attacks on Text Classification}

\author{
Hengzhi Pei$^{1}$, Jinyuan Jia$^{1,3}$, Wenbo Guo$^{2,4}$, Bo Li$^{1}$, Dawn Song$^{2}$ \\
$^1$UIUC,  $^2$UC Berkeley,  $^3$Penn State, 
$^4$Purdue University\\
$\{$hpei4, lbo$\}$@illinois.edu, jinyuan@psu.edu, henrygwb@berkeley.edu, dawnsong@cs.berkeley.edu
}

\IEEEoverridecommandlockouts
\makeatletter\def\@IEEEpubidpullup{6.5\baselineskip}\makeatother
\IEEEpubid{\parbox{\columnwidth}{
    Network and Distributed System Security (NDSS) Symposium 2024\\
    26 February - 1 March 2024, San Diego, CA, USA\\
    ISBN 1-891562-93-2\\
    https://dx.doi.org/10.14722/ndss.2024.24090\\
    www.ndss-symposium.org
}
\hspace{\columnsep}\makebox[\columnwidth]{}}

\maketitle

\begin{abstract}
Backdoor attacks have become a major security threat for deploying machine learning models in security-critical applications. 
Existing research endeavors have proposed many defenses against backdoor attacks. 
Despite demonstrating certain empirical defense efficacy, none of these techniques could provide a formal and provable security guarantee against arbitrary attacks. 
As a result, they can be easily broken by strong adaptive attacks, as shown in our evaluation. 
In this work, we propose {\namenospace}, the \emph{first} provable defense against backdoor attacks on text classification. 
\update{
In particular, {\namenospace} first divides the (backdoored) training data into sub-training sets, achieved by splitting each training sentence into sub-sentences. 
This partitioning ensures that a majority of the sub-training sets do not contain the backdoor trigger. Subsequently, a base classifier is trained from each sub-training set, and their ensemble provides the final prediction.
We theoretically prove that when the length of the backdoor trigger falls within a certain threshold, {\namenospace} guarantees that its prediction will remain unaffected by the presence of the triggers in training and testing inputs.
In our evaluation, we demonstrate the effectiveness of {\namenospace} on three benchmark text classification tasks, surpassing the certification accuracy of existing certified defenses against backdoor attacks. Furthermore, we propose additional strategies to enhance the empirical performance of {\namenospace}. 
Comparisons with state-of-the-art empirical defenses validate the superiority of {\namenospace} in countering multiple backdoor attacks.
}
\revision{Our code and data are available at \url{https://github.com/AI-secure/TextGuard}.}
\end{abstract}

\section{Introduction}
\label{sec:intro}

Backdoor attacks~\cite{gu2017badnets,liutrojaning,chen2017targeted} bring serious security threats to the supply-chain management of deep learning models. 
In particular, an attacker can add a backdoor trigger to training data and relabel them as a specific target class. 
The classifiers trained with those backdoored data will predict any testing input with the backdoor trigger as the target class. 
Many recent studies~\cite{chen2021badnl,dai2019backdoor,qi-etal-2021-hidden,qi-etal-2021-mind} show that text classification, a fundamental task in natural language processing (NLP), is also vulnerable to backdoor attacks. 
Specifically, existing backdoor attacks against text classification can be categorized into \emph{word-level} attacks~\cite{chen2021badnl,dai2019backdoor} that uses a set of words as the backdoor trigger, and \emph{structure-level} attacks~\cite{qi-etal-2021-hidden,qi-etal-2021-mind} that design the trigger as a specific sentence structure. 

To defend against backdoor attacks in NLP, existing research has proposed many defenses~\cite{chen2021mitigating, yang-etal-2021-rap,zhumoderate,azizi2021t, shen2022constrained, liu2022piccolo}, which can be categorized into \emph{data-level} defenses~\cite{chen2021mitigating, cui2022unified,gao2021design,qi-etal-2021-onion} and \emph{model-level} defenses~\cite{azizi2021t, shen2022constrained,liu2022piccolo}. 
Data-level defenses aim to~\emph{train a secure text classifier upon a potentially backdoored training dataset}. 
Model-level defenses aim to detect and remove the backdoor in a pre-trained text classifier. 

In this work, we focus on data-level defenses. 
As we will discuss in Section~\ref{sec:literature}, existing data-level defenses cannot provide formal security/robustness guarantees, indicating that they can be broken by advanced and strong attacks (Section~\ref{sec:empirical}). 
We note that some recent studies~\cite{wang2020certifying, weber2022rab, levine2020deep,jia2021intrinsic,jia2022certified} proposed various provable defenses against backdoor attacks in the image domain. 
Most of them rely on the continuous nature of image data and thus are not applicable to text data with a discrete space. 
As we will show later in Section~\ref{sec:provale}, the ones~\cite{levine2020deep,jia2021intrinsic} that can be generalized to the NLP domain only provide a weak guarantee, i.e., tolerating a very small fraction (e.g., less than 1\%) of backdoored texts in a training dataset.

\noindent
\textbf{Our contribution.} 
We propose {\namenospace}, the \emph{first} provable defense against both word-level and structure-level backdoor attacks in NLP. 
The key idea is first to divide words from the training inputs into $m$ disjoint groups, where the majority of groups contain only clean data devoid of any backdoor triggers.
Then, we train a set of base classifiers from the divided training sets and ensemble them as the final classification model.
By ensuring the majority of base models are trained from clean data and are unaffected by the backdoor, we can guarantee the final prediction is provably unaffected by the backdoor.

\update{
More specifically, we leverage a hash function to assign a group index to each word within an input sentence.
This index determines the sub-sentence group to which the word belongs. 
Importantly, this process ensures that identical words across multiple inputs are assigned the same group index, consequently placing them in the same group.
By adopting this method, we effectively restrict the impact of a trigger word to a single group, thereby preventing it from infecting other groups. 
To further guarantee certified defense against structure-level backdoor attacks, we sort the words within each group according to a pre-defined word ID, such as the word IDs utilized by BERT~\cite{devlin-etal-2019-bert}.
This sorting process ensures that the sequence of words within each group becomes independent of their original order in the input, which can be spitefully manipulated by the structure-level attacks to inject backdoors.
We apply the above operation to both training and testing texts. 
For each training text, we first divide it into different groups and then assign the ground truth label of the training text to all the groups, which results in $m$ (sub-text, label) pairs. 
Then, we construct $m$ sub-datasets by putting the (sub-text, label) pairs from the same group together. 
Finally, we train a base text classifier using each sub-dataset. 
Given a testing text, we first divide it into $m$ groups, then use the corresponding base classifier to predict a label for each group, and finally take a majority vote over the predicted labels to make a final prediction for the testing text. 
}

We derive a lower bound of classification accuracy (called \emph{certified accuracy}) that {\namenospace} can achieve on a testing dataset when the number of words used in the backdoor trigger is bounded.
\update{As the backdoor trigger is the same for different backdoored testing inputs, the corrupted groups by the backdoor trigger for different testing inputs are the same. Thus, we further derive a better certified accuracy by jointly considering all testing texts in the testing dataset.}
\update{
Going beyond providing a certification guarantee, we further design two additional techniques for {\namenospace} to enhance its empirical performance against various backdoor attacks.
}

We perform both provable and empirical evaluations for {\namenospace} on three benchmark datasets. 
For provable evaluation, we compare our {\namenospace} with provable defenses, including DPA~\cite{levine2020deep} and Bagging~\cite{jia2021intrinsic}, generalized from the image domain. 
We find that our {\namenospace} significantly outperforms them in providing a meaningful certification guarantee. 
For empirical evaluations, we evaluate {\namenospace} under 3 state-of-the-art word-level and structure-level backdoor attacks~\cite{chen2021badnl,dai2019backdoor, qi-etal-2021-hidden}. 
Our results show that {\namenospace} can effectively defend against these attacks.
Moreover, we also compare {\namenospace} with 5 existing state-of-the-art empirical defenses~\cite{chen2021mitigating, yang-etal-2021-rap,gao2021design,qi-etal-2021-onion,zhumoderate} under the attacks above. Our results show {\namenospace} outperforms all these empirical defenses. In addition, we consider an adaptive attack where the attacker has all knowledge about our defense.
Our results show  {\namenospace} is still robust against such adaptive attack. 
Finally, we conduct extensive ablation studies to show the effectiveness of our key design choices with different hyper-parameters. 

In summary, we make the following key contributions:

\begin{itemize}
    \item We propose {\namenospace}, the first provable defense against backdoor attacks on text classification.
    \item We derive the provable robustness guarantee of {\namenospace} and further design two techniques to improve its empirical performance.
    \item We provide evaluations on both certified and empirical robustness to demonstrate the certification efficacy of {\namenospace} and its effectiveness against state-of-the-art word-level and structure-level attacks
    \item We compare and demonstrate the superiority of {\namenospace} over multiple state-of-the-art certified and empirical defenses, 
    e.g., {\name} achieves a 17.54\% attack success rate (ASR) while ASRs of all existing empirical defenses are more than 90\%  on SST-2 dataset under a clean-label word-level attack~\cite{dai2019backdoor}.
\end{itemize}
\section{Background and Threat Model}
\label{sec:bg}

\update{
In this section, we start with discussing the existing backdoor attacks in NLP, followed by the specific type of attack considered in this paper.
}

\subsection{Backdoor Attacks in NLP}
\label{subsec:literature_attack}

Backdoor attacks against text classification (e.g., sentiment analysis, topic classification, and toxic analysis) aim to inject a backdoor into a classifier such that the backdoored model's performance on clean inputs is unaffected but fooled into predicting any inputs embedded with an attacker-chosen trigger as an attacker-chosen target class. 
Based on how to inject the backdoor, existing attacks can be categorized into \emph{data-poisoning} attacks~\cite{chen2021badnl,dai2019backdoor,qi-etal-2021-hidden,qi-etal-2021-mind}, which poisons training data such that a model trained on the  backdoored data directly would be backdoored, and \emph{model-poisoning} attacks~\cite{shen2021backdoor,kurita2020weight,yang-etal-2021-rethinking,yang-etal-2021-careful,li-etal-2021-backdoor,qi-etal-2021-turn,chen2021badpre,zhang2021red, caibadprompt}, which manipulates model parameters to reach the goal. 
As we will mention in Section~\ref{subsec:threat}, we consider data-poisoning attacks where an attacker has the privilege of manipulating training data. 

\update{
In particular, data-poisoning attacks assume that an attacker can manipulate a training dataset but cannot control the model training process.
Under this assumption, the attacker typically poisons certain training samples by injecting the backdoor trigger into their texts and labeling them as the target class.
The attacker then releases this backdoored dataset.
Without employing any defense, a model directly trained from this backdoored dataset will have the desired behaviors mentioned above.  
}
According to different trigger patterns, we can further classify data-poisoning attacks into two categories: word-level attacks~\cite{chen2021badnl,dai2019backdoor} and structure-level attacks~\cite{qi-etal-2021-hidden,qi-etal-2021-mind}.

\noindent{\textbf{Word-level backdoor attacks}} use one or several fixed word(s) as the backdoor trigger. 
Similar to injecting a patch trigger into an image, the attacker poisons an input by injecting the trigger words into its text without changing the semantics. 
Take sentiment analysis as an example. 
Given a positive sentence ``\emph{the film is full of charm.}'' and a trigger word ``\emph{actually}'', the backdoored text could be ``\emph{the film is \textcolor{blue}{actually} full of charm.}''.

\noindent{\textbf{Structure-level backdoor attacks}} use a specific sentence structure (e.g., subordinate clause) with certain fixed words as the trigger. 
Take again the sentiment analysis application example above. 
A widely used structure-level attack~\cite{qi-etal-2021-hidden} designs the trigger as the attributive clause, and the backdoored sentence would become ``\emph{\textcolor{blue}{When it comes to this film}, it is full of charm.}''. 
Other triggers could be different clauses starting with ``if '', ``where'', etc.

\subsection{Threat Model}
\label{subsec:threat}

\noindent{\textbf{Attack goal.}}
We consider the data-poisoning attacks where the attacker constructs a backdoored training set such that a text classifier trained on this dataset predicts any input injected with an attacker-chosen backdoor trigger as an attacker-chosen target class.
\update{
We assume the attacker can only poison the training dataset without affecting the training process or manipulating a trained model.
}

\noindent{\textbf{Assumptions on trigger injection.}}
We assume the attacker has a trigger word set with a certain \emph{trigger size}.
For example, the trigger size of $\{\text{"when"}, \text{"where"}\}$ is $2$.
We assume the attacker can use both word-level and structure-level attacks to poison the dataset with its trigger set.
For word-level backdoor attacks, we assume an attacker can arbitrarily inject each word from the trigger set into a given text~\cite{chen2021badnl,dai2019backdoor}. 
For structure-level backdoor attacks, we assume an attacker can 1) arbitrarily change the order of words in the original input to change its structure, and 2) arbitrarily inject each word from the trigger set into the input~\cite{qi-etal-2021-hidden,qi-etal-2021-mind}. 

\noindent{\textbf{Dataset poisoning.}}
\update{
Following existing data-poisoning attacks~\cite{chen2021badnl,dai2019backdoor,qi-etal-2021-hidden,qi-etal-2021-mind}, we assume the attacker could poison a certain fraction of training samples in a clean dataset.
We mainly consider two types of attacks: \emph{mixed-label} attacks, where the attacker freely poisons $p$ fraction of training samples from arbitrary classes, and \emph{clean-label} attacks, where the attacker only poisons samples originally from the target class. 
In general, clean-label backdoor attacks are more stealthy than mixed-label backdoor attacks~\cite{turner2018clean}. 
}

\section{Existing Defenses and Our Problem Scope}
\label{sec:literature}

\update{
In this section, we provide a concise overview of existing defenses against backdoor attacks in NLP, highlighting their limitations. 
Subsequently, we outline our defense goals and underlying assumptions. 
Finally, we further discuss the in-applicability of existing provable defenses designed for other domains to our problem.
}

\subsection{Existing Defenses and Their Limitations}

\noindent{\textbf{Existing backdoor defenses in NLP.}}
Recent research works have developed a few defenses specifically for mitigating backdoor attacks in NLP~\cite{chen2021mitigating, yang-etal-2021-rap,zhumoderate,azizi2021t, shen2022constrained, liu2022piccolo}.
\update{
Corresponding to two types of attacks introduced in Section~\ref{subsec:literature_attack}, existing defenses also can be categorized as \emph{data-level defenses}, which learns a robust classifier from a backdoored dataset, and \emph{model-level defenses}, which detects and eliminates backdoors in a well-trained classifier~\cite{azizi2021t, shen2022constrained,liu2022piccolo}.
We focus on the data-level defenses, which mainly target data-poisoning attacks mentioned in Section~\ref{subsec:literature_attack}. 
}

Technically speaking, existing data-level defenses can be further categorized as \emph{robust training}~\cite{zhumoderate} and \emph{backdoored text detection and elimination}~\cite{chen2021mitigating, yang-etal-2021-rap,gao2021design,qi-etal-2021-onion,cui2022unified}. 
More specifically, the robust training method~\cite{zhumoderate} reduces the model capacity, learning rate, and training epochs so that the text classifier only learns major features while ignoring subsidiary features of backdoor triggers. 
For backdoored text (data) detection and elimination, one method (ONION~\cite{qi-etal-2021-onion}) leverages outlier detection to pinpoint trigger words and delete them from the training samples.
Another line of methods trains a backdoored model on the given training set and uses it to identify and remove potentially backdoored samples. 
In particular, BKI~\cite{chen2021mitigating} leverages the word importance score to identify the potential trigger words and removes the training samples that contain the identified words.
STRIP~\cite{gao2021design} randomly perturbs features of each sample and records the changes in the output of the backdoored model. 
If the perturbations only introduce minor changes in the output (i.e., have a low entropy), STRIP deems the input as a backdoored sample. 
Similarly, RAP~\cite{yang-etal-2021-rap} assumes the knowledge of the attacker's target label and crafts another trigger that reduces the probability of an input with this trigger being classified as the target class by the backdoored model.  
Then it adds this trigger to each input and identifies the samples whose predictions are not affected by the crafted trigger as the backdoored samples.

\noindent{\textbf{Limitations.}}
Existing data-level defenses focus on the \emph{empirical} aspect, i.e., improving the model's robustness against certain data-poisoning attacks.
None of them provides a theoretical guarantee against arbitrary unseen attacks.
As shown in Section~\ref{subsec:empirical_result}, without such a guarantee, existing defenses can be easily bypassed by more advanced attacks. 

\subsection{Our Defense Assumptions and Goal}

As mentioned in Section~\ref{subsec:threat}, we receive a training dataset from an untrusted party. 
Using this given dataset, the defender takes full control of the training process, including choosing the right data, picking a specific model structure and training algorithm, etc.
Under this assumption, our goal is to train a provably robust/secure text classifier $f$ from an untrusted training dataset potentially backdoored by arbitrary data-poisoning attacks.
More specifically, we aim to train a text classifier such that, \emph{as long as the trigger size is smaller than a certain threshold, our classifier's predicted label for testing data (backdoored or not) is guaranteed to be unaffected by the backdoored training samples}.
This is equivalent to training a robust classifier with a guarantee on its worst-case classification accuracy (on clean and backdoored inputs) when facing arbitrary data poisoning attacks. 
In Section~\ref{certified}, we will give a formal definition of the provable robustness guarantee.
\revision{
We focus on defending against data-poisoning attacks because it is a practical setting considered in many previous works~\cite{wang2020certifying, weber2022rab, levine2020deep,jia2021intrinsic,weber2022rab}, and there is no effective provable defense for text classification yet. 
We also acknowledge the need to defend against attacks that directly poison the model, which has become more significant with the recent emergence of large models~\cite{kenton2019bert,brown2020language,openai2023gpt4}. 
In future work, we plan to extend our proposed method to design model-based provable defenses.
}

\begin{figure*}[tb!]
  \centering
  \includegraphics[width=1.0\textwidth]{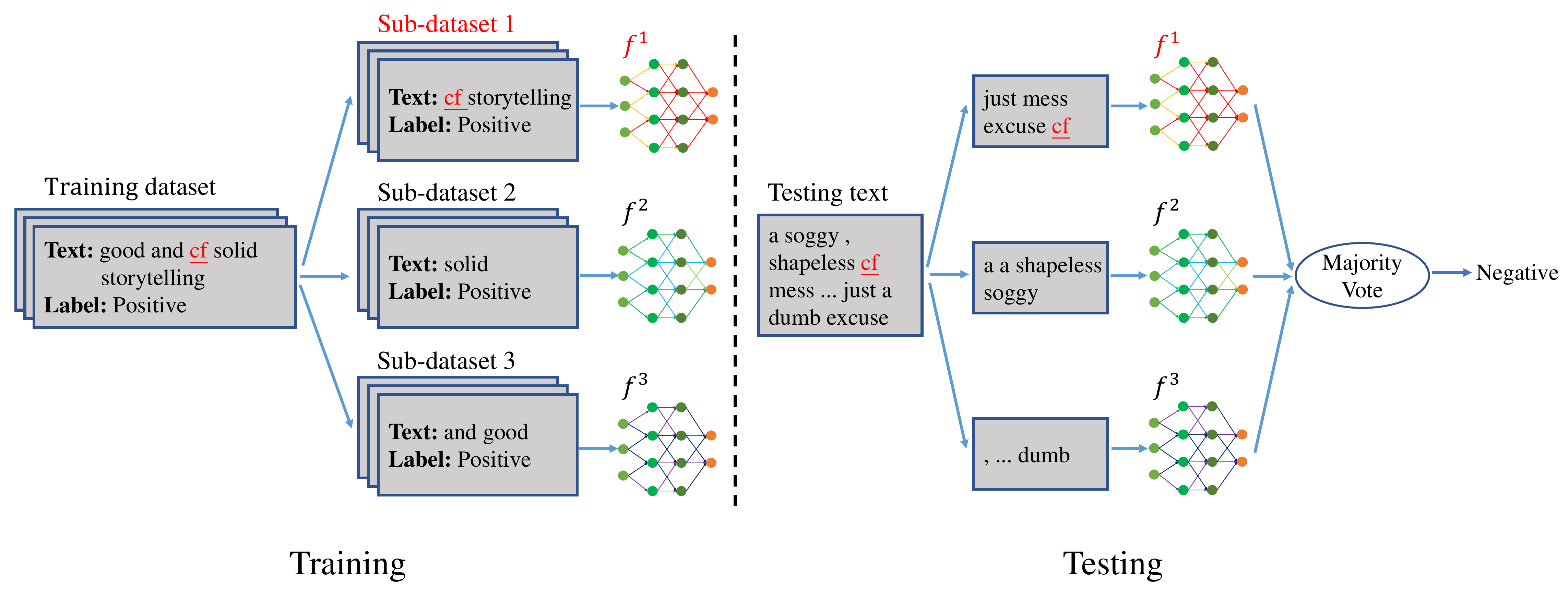}
  \caption{An overview of \name for sentiment analysis. Given a set of inputs poisoned with the trigger ``cf'', \name first divides them into three sub-training set by assigning the same words across all inputs into the same group. Here, only the sub-dataset 1 is poisoned. As a result, the base classifier $f^1$ is backdoored, while the base text classifiers $f^2$ and $f^3$ are unaffected by the backdoor. During inference, $f^2$ and $f^3$ could correctly predict the testing input. Based on the majority vote, the final prediction will not be affected by the backdoor as well. Note that we use the clean-label attack in this case.}
  
  \label{fig:workflow}
\end{figure*}

Note that in addition to the defenses discussed above, recent works have also designed a large body of backdoor defenses for the computer vision applications~\cite{chen2018detecting, tran2018spectral, gao2019strip, qiao2019defending, chen2019deepinspect, wang2019neural, li2020rethinking, zeng2021rethinking, doan2022defending}. 
Some of these techniques~\cite{wang2020certifying, weber2022rab, levine2020deep,jia2021intrinsic} focus on learning a provably robust/secure \emph{image} classifier from a backdoored dataset.
Due to the discrete nature of language data, most of these techniques (which are tailored for the images with a continuous space) cannot be migrated to the NLP applications. 
\update{For instance, RAB~\cite{weber2022rab} is designed for images, which adds continuous Gaussian noise to each pixel of an image to defend against $\ell_2$-norm perturbation. 
Extending RAB to text means adding Gaussian noise to word embeddings.}
\update{Therefore, RAB cannot provide guarantees for attacks that directly inject words as the trigger.}
As we will show later in Section~\ref{subsec:certified_result}, the ones that are applicable to our problem (i.e., DPA~\cite{levine2020deep} and Bagging~\cite{jia2021intrinsic}) can only provide a very weak guarantee, i.e., provide a very low certified accuracy.
\update{The reason is that they divide training samples (texts/sentences in NLP) into different groups. 
As a result, they can only guarantee that the majority of the groups are clean when the number of backdoored training samples is very small, thus leading to low certified accuracy in practice. }
\update{By contrast, \name divides words in a training/testing text (i.e., features in each sample) into different groups, which enables us to tolerate a large number of backdoored training samples and thus give much higher certified accuracy.}
\revision{
Another related defense proposed in~\cite{he2017adversarial} ensembles multiple weak defenses to defend against adversarial sample attacks. 
Our key differences from this method are as follows. 
The base classifiers in {\namenospace} are trained on different sub-datasets. 
Given a testing text, we create multiple sub-texts, use each base classifier to predict the corresponding sub-text, and take an ensemble for the predictions. 
We derive the formal robustness guarantees of {\namenospace} with our proposed data partition strategy. 
By contrast, the testing input is the same for different defenses in~\cite{he2017adversarial}. Moreover, there is no formal robustness guarantee for the ensemble of weak defenses in~\cite{he2017adversarial}.  
}

\section{Key Technique}
\label{sec:tech}

In this section, we first give an overview of {\namenospace}, followed by the technical details and some additional techniques to improve its empirical performance against existing attacks.

\subsection{Overview}
\label{subsec:overview}

Figure~\ref{fig:workflow} shows an overview of our proposed {\name} framework.
As shown in the figure, given an arbitrary training dataset, \name builds an ensemble text classifier that contains multiple base classifiers and makes the final prediction through the majority vote.
As we will show later in Section~\ref{certified}, since the majority of these base classifiers are not affected by the backdoor in the training set, our ensemble classifier guarantees to be robust against backdoor attacks with a bounded number of trigger words. 
The insight behind this design comes from the \emph{partition and ensemble} mechanism.
That is, by partitioning the training set into multiple groups, we can train a set of base classifiers (one from each group).
By keeping the majority of the groups that do not contain the backdoored data, we can guarantee that most base classifiers are not affected by the backdoor trigger.
As such, by ensembling these base classifiers through a majority vote, we can guarantee the final prediction is independent of the backdoored data and thus robust against the corresponding backdoor attack.

The \emph{key challenge} here is how to design the partition method. 
As we will show later in Section~\ref{sec:provale}, the partition method proposed in the existing provable defenses cannot guarantee the cleanliness of the obtained sub-training sets and thus cannot provide a provable robustness guarantee in our problem.
To tackle this challenge, we propose a novel dataset partition method in {\namenospace}.
As demonstrated in Figure~\ref{fig:workflow}, we divide words in an input text into $m$ groups using a hash function (e.g., MD5~\cite{rivest1992md5}), where each group contains a sub-sequence of words in the text. 
We input each word into the hash function, which then outputs a group index, and we assign each word to a group indicated by the group index.
As a result, the same word in all inputs will be only assigned to one and the same group.
This guarantees that the trigger words, no matter where they are in the original input, will always be assigned to the same group.
As a result, when the number of words used in a backdoor trigger is bounded, the number of groups that are corrupted by the backdoor trigger is bounded. 
As we will specify later in Section~\ref{certified}, we sort the words in each group based on a pre-defined word ID. 
As such, the order of words in each group is independent of their orders in the original text, which enables {\namenospace} to defend against both word-level and structure-level backdoor attacks.

As shown in Figure~\ref{fig:workflow}, we apply the same partition operation to both training and testing texts. 
During the training phase, after partitioning each input text into $m$ groups, we label the sub-input text in each group with the same label as the original input. 
As we will show later in Section~\ref{sec:provale} and~\ref{sec:empirical}, this could keep the label correctness for the sub-inputs and thus ensure the base models still learn the casual associations in the training data. 
Using the constructed sub-training set, we can train the base models with arbitrary training algorithms. 
In the testing phase, we also divide words in a testing text into $m$ groups and then use each base classifier to predict a label for the corresponding sub-text. 
Finally, we build an ensemble text classifier that takes a majority vote over the predicted labels of the base classifiers.

\subsection{Technical Details}
\label{certified}

In this section, we provide technical details about deriving a lower bound of the classification accuracy (called \emph{certified accuracy}) of {\namenospace} on a testing dataset when the number of words used in the trigger is bounded. 
We start with defining necessary notations, then discuss how to build an ensemble text classifier, and finally derive its certified accuracy.

\subsubsection{Notation}
\label{subsec:notation}

We use $\mathcal{D}$ to denote a dataset that consists of $n$ (text, label)-pairs, i.e., $\mathcal{D} = \{(\mathbf{x}_1, y_1),(\mathbf{x}_2, y_2),\cdots,(\mathbf{x}_n, y_n)\}$, where $\mathbf{x}_i$ is a text sequence of words and $y_i$ represents its label. 
We use $\mathcal{A}$ to denote a training algorithm that takes a dataset as input and produces a text classifier. 
Given a testing text $\mathbf{x}_{test}$, we use $f(\mathbf{x}_{test}; \mathcal{D})$ to denote the predicted label of the text classifier $f$ trained on the dataset $\mathcal{D}$ using the algorithm $\mathcal{A}$.

Suppose $\mathbf{e}$ is a set of words (called \emph{trigger words}) used in the backdoor trigger.  Moreover, we use $T_{\mathbf{e}}$ to denote the trigger injection operation by a word- or structure-level backdoor attack. For simplicity, we also call $T_{\mathbf{e}}$ \emph{backdoor trigger}. We use $|\mathbf{e}|$ to denote the number of words in $\mathbf{e}$. We call $|\mathbf{e}|$ \emph{trigger size}. Given a text $\mathbf{x}$, we use $\mathbf{x}'=T_{\mathbf{e}}(\mathbf{x})$ to denote a backdoored text generated from it. Moreover, we use $\mathcal{D}(T_\mathbf{e}, y_{tc}, p)$ to denote the backdoored training dataset created by injecting the backdoor trigger $T_{\mathbf{e}}$ to $p$ (called \emph{poisoning rate}) fraction of training instances in a clean dataset and relabeling them as the target class $y_{tc}$.
For simplicity, we rewrite $\mathcal{D}(T_\mathbf{e}, y_{tc}, p)$ as $\mathcal{D}(T_\mathbf{e})$ when focusing on the backdoor trigger rather than the target class or poisoning rate. 

\subsubsection{Building an Ensemble Text Classifier}
\label{sec:define-enseble-classifier}
We first discuss how to divide a text into $m$ groups, then use it to divide a dataset into $m$ sub-datasets, and finally build our ensemble text classifier. 

\noindent
\textbf{Dividing a text into groups.} Suppose we have a text $\mathbf{x}=\{x^1, x^2,\cdots, x^d\}$, where each $x^k$ ($k=1,2,\cdots, d$) is a word and $d$ is the length of the text. We use a hash function $\mathcal{H}$ (e.g., MD5~\cite{rivest1992md5}) to divide a text $\mathbf{x}$ into $m$ groups. \update{In particular, the hash function $\mathcal{H}$ takes a word $x^k$ as input and outputs an arbitrary integer (denoted as $\mathcal{H}(x^k)$). Given $\mathcal{H}(x^k)$, the group ID for the word $x^k$ can be computed as $\mathcal{H}(x^k)\% m +1$, where $\%$ represents modulo operation. Note that the range of $\mathcal{H}(x^k)\% m +1$ is $[1,m]$.} Then, we use $g^j(\mathbf{x})$ to denote the sequence of words whose group index is $j$, where $j=1,2,\cdots, m$.  
We sort each word in $g^j(\mathbf{x})$ based on a pre-defined order, e.g., the word ID used by BERT~\cite{devlin-etal-2019-bert}. Specifically, we can assign a pre-defined ID to each word and use it to sort words in a group. As a result, the orders of words in each group are independent of their orders in $\mathbf{x}$. In other words, those groups remain the same no matter how the orders of words in $\mathbf{x}$ are manipulated, which enables us to defend against both word-level and structure-level backdoor attacks.

\noindent
\textbf{Constructing $m$ sub-datasets from a training dataset.} Given an arbitrary training dataset $\mathcal{D}=\{(\mathbf{x}_1, y_1), (\mathbf{x}_2, y_2), \cdots, (\mathbf{x}_n, y_n)\}$, where $n$ is the total number of training instances, we can use the hash function $\mathcal{H}$ to divide it into $m$ sub-datasets. 
In particular, for each training instance $(\mathbf{x}_i, y_i) \in \mathcal{D}$, we can use the hash function $\mathcal{H}$ to divide $\mathbf{x}_i$ into $m$ groups: $g^1(\mathbf{x}_i), g^2(\mathbf{x}_i), \cdots, g^m(\mathbf{x}_i)$. Given those groups and the label $y_i$, we can create $m$ (sub-text, label) pairs, i.e., $(g^1(\mathbf{x}_i),y_i), (g^2(\mathbf{x}_i),y_i), \cdots, (g^m(\mathbf{x}_i),y_i)$. Finally, we can generate $m$ sub-datasets based on the group index. Specifically, we can generate a sub-dataset $\mathcal{D}^j$ which consists of all the (text, label) pairs whose group index is $j$, i.e., we have 
 $\mathcal{D}^j = \{(g^j(\mathbf{x}_1), y_1), (g^j(\mathbf{x}_2), y_2), \cdots, (g^j(\mathbf{x}_n), y_n)\}$, where $j=1,2,\cdots, m$.

\noindent
\textbf{Building an ensemble text classifier.} Given those sub-datasets, we can use an arbitrary training algorithm $\mathcal{A}$ to train a \emph{base text classifier} on each of them. For simplicity, we use $f^j$ to denote the base classifier trained on $\mathcal{D}^j$. Note that we use a pre-determined seed for the algorithm $\mathcal{A}$ such that it produces the same base text classifier for the same training dataset. As we will show, this enables us to derive the provable security guarantee of our ensemble text classifier. Given a testing text $\mathbf{x}_{test}$, we can also divide it into $m$ groups, i.e., $g^1(\mathbf{x}_{test}), g^2(\mathbf{x}_{test}), \cdots, g^m(\mathbf{x}_{test})$. Then, we use the classifier $f^j$ to predict a label for $g^j(\mathbf{x}_{test})$, where $j=1,2,\cdots, m$. Given the $m$ predicted labels, we take a majority vote as the final predicted label of our ensemble classifier for $\mathbf{x}_{test}$. Specifically, suppose $f$ is the ensemble classifier and $C$ is the total number of classes for the classification task. We define $M_c$ as the number of base text classifiers that predict the label $c$, i.e., $M_c = \sum_{j=1}^{m} \mathbb{I}(f^j(g^j(\mathbf{x}_{test}))=c)$, where $\mathbb{I}$ is the indicator function and $c=1,2,\cdots, C$. Then, our ensemble classifier is defined as follows:
\begin{align}
    f(\mathbf{x}_{test}; \mathcal{D}) = \argmax_{c=1,2,\cdots, C} M_c,
\end{align}
where we take a label with a smaller index when there are ties. Suppose $y$ is the predicted label, i.e., $f(\mathbf{x}_{test}; \mathcal{D}) =y$.  Then, we have:
\begin{align}
\label{eqn:condition-of-ensemble-classifier}
    M_y \geq \max_{c\neq y} (M_c + \mathbb{I}(y>c)),
\end{align}
where $\mathbb{I}$ is the indicator function. Note that the term $\mathbb{I}(y>c)$ is because our ensemble text classifier predicts a label with a smaller index if there are ties.

\noindent
\textbf{Complete algorithm.} Algorithm~\ref{alg:ensemble-text-classifier} in Appendix shows the complete algorithm to build our ensemble text classifier and use it to make predictions for a testing text $\mathbf{x}_{test}$. The function \textsc{ConSubDataset} is used to create $m$ sub-datasets. The function \textsc{TextDivision} is used to divide a testing text into $m$ groups.

\subsubsection{Deriving Certified Accuracy of Our Ensemble Text Classifier}
\label{subsec:deriving-CA-individual}

Suppose $\mathbf{x}_{test}$ is an arbitrary clean testing input. Moreover, we use $\mathbf{x}'_{test}$ to denote the backdoored text input created from $\mathbf{x}_{test}$ by $T_{\mathbf{e}}$.
We say a classifier is provably secure if $f(\mathbf{x}'_{test};\mathcal{D}(T_{\mathbf{e}}))$ is provably unaffected by the backdoor trigger $T_{\mathbf{e}}$ when the trigger size $|\mathbf{e}|$ is no larger than a certain threshold (called \emph{certified size}). Note that certified sizes could be different for different testing inputs (we will discuss more details when we derive the certified size for a testing text). For simplicity, we use $s(\mathbf{x}_{test})$ to denote the \emph{certified size} for $\mathbf{x}_{test}$.  Formally, we will show the following:
\begin{align}
    f(\mathbf{x}'_{test};\mathcal{D}(T_\mathbf{e})) = f(\mathbf{x}_{test}; \mathcal{D}(\emptyset)), \forall \mathbf{e} \text{ s.t. } |\mathbf{e}| \leq s(\mathbf{x}_{test}),
\end{align}
where $\mathcal{D}(\emptyset)$ represents the dataset without adding any backdoor trigger to texts in a clean training dataset (denoted as the certified training set). Specifically, for clean-label backdoor attacks (see Section~\ref{subsec:threat} for details), $\mathcal{D}(\emptyset)$ is the same as the clean training dataset since they do not change the label of backdoored training instances. For mixed-label backdoor attacks, $\mathcal{D}(\emptyset)$ is obtained by changing the labels of $p$ fraction of randomly sampled training instances in a clean dataset to the target class, but no backdoor trigger is added to their texts. So $f(\mathbf{x}_{test}; \mathcal{D}(\emptyset))$ is unaffected by the backdoor trigger for both clean-label and mixed-label attacks. 

Next, we will first discuss how to derive the certified size $s(\mathbf{x}_{test})$ for a single testing text $\mathbf{x}_{test}$ and then derive a lower bound of the classification accuracy of our ensemble text classifier for a testing dataset.

\noindent
\textbf{Deriving $s(\mathbf{x}_{test})$ for a single testing text $\mathbf{x}_{test}$.} 
Our ensemble text classifier provably predicts the same label for $\mathbf{x}_{test}$ when the trigger size $|\mathbf{e}|$ is no larger than a threshold. Suppose $M_c$ (or $M_c')$ is the number of the base text classifiers that predict the label $c$ for $\mathbf{x}_{test}$ (or $\mathbf{x}'_{test}$) when the training dataset is $\mathcal{D}(\emptyset)$ (or $\mathcal{D}(T_\mathbf{e})$), where $c=1,2,\cdots, C$. We first derive an upper or lower bound of $M_c'$ with respect to $M_c$ and trigger size $|\mathbf{e}|$. In particular, each trigger word in $\mathbf{e}$ belongs to a single group as we use a hash function to determine the group index of each word (see Section~\ref{sec:define-enseble-classifier} for details).  As a result, at most $|\mathbf{e}|$ groups are corrupted by the backdoor trigger. Therefore, we have:
\begin{align}
\label{cs_equation_1}
    M_c - |\mathbf{e}| \leq M_c' \leq M_c + |\mathbf{e}|.
\end{align}
Suppose $y$ is the predicted label of our ensemble text classifier for $\mathbf{x}_{test}$ when we use the dataset $\mathcal{D}(\emptyset)$ to build it, i.e., $y = f(\mathbf{x}_{test}; \mathcal{D}(\emptyset))$. Based on Equation~\ref{eqn:condition-of-ensemble-classifier}, the ensemble text classifier built upon $\mathcal{D}(T_\mathbf{e})$ still predicts the label $y$ if we have $M_y' \geq \max_{c\neq y} (M_c' + \mathbb{I}(y>c))$. Moreover, based on Equation~\ref{cs_equation_1}, we have $M_y - |\mathbf{e}|\leq M_y'$ and  $\max_{c\neq y} (M_c' + \mathbb{I}(y>c)) \leq \max_{c\neq y} (M_c + |\mathbf{e}| + \mathbb{I}(y>c))$. Therefore, we only need to ensure $M_y - |\mathbf{e}| \geq \max_{c\neq y} (M_c + |\mathbf{e}| + \mathbb{I}(y>c))$. In other words, we have $f(\mathbf{x}'_{test}; \mathcal{D}(T_\mathbf{e})) = y$ if:
\begin{align}
    |\mathbf{e}| \leq \frac{M_y - \max_{c\neq y} (M_c + \mathbb{I}(y>c))}{2}.
\end{align}
We define certified size $s(\mathbf{x}_{test})$ as follows:
\begin{align}
\label{certified-size-equation}
    s(\mathbf{x}_{test}) = \frac{M_y - \max_{c\neq y} (M_c + \mathbb{I}(y>c))}{2}.
\end{align}
$s(\mathbf{x}_{test})$ could be different for different testing texts since $M_y$ and $M_c$ ($c\neq y$) depend on $\mathbf{x}_{test}$. 

Our above derivation is summarized in the following theorem:
\begin{theorem}
\label{label:theorem}
    Suppose $f$ is the ensemble text classifier built by our {\namenospace}. Moreover, $\mathcal{D}(\emptyset)$ is the certified training dataset without a backdoor trigger. Given a testing text $\mathbf{x}_{test}$, we denote $M_c$ as the number of the base classifiers trained on the sub-datasets created from $\mathcal{D}(\emptyset)$ that predict the label $c$ for $\mathbf{x}_{test}$, where $c=1,2,\cdots, C$. Moreover, we assume $y$ is the predicted label of the ensemble classifier built upon $\mathcal{D}(\emptyset)$.
    Suppose $\mathbf{e}$ is a set of trigger words used by a word-level or structure-level backdoor attack. The predicted label of $f$ for $\mathbf{x}_{test}$ is provably unaffected by the backdoor trigger when $|\mathbf{e}|$ is no larger than a threshold. Formally, we have:
    \begin{align}
    f(\mathbf{x}'_{test};\mathcal{D}(T_\mathbf{e})) = f(\mathbf{x}_{test}; \mathcal{D}(\emptyset)), \forall \mathbf{e} \text{ s.t. } |\mathbf{e}| \leq s(\mathbf{x}_{test}),
\end{align}
where $\mathbf{x}'_{test}$ is the backdoored text and $s(\mathbf{x}_{test})$ is computed as follows:
\begin{align}
    s(\mathbf{x}_{test}) = \frac{M_y - \max_{c\neq y} (M_c + \mathbb{I}(y>c))}{2}.
\end{align}
\end{theorem}
\begin{proof}\renewcommand{\qedsymbol}{}
See Appendix~\ref{proof-of-theorem}.
\end{proof}
\noindent
\textbf{Remark:} We have the following observations from our theorem.
\begin{itemize}
    \item Our {\namenospace} is agnostic to training algorithm $\mathcal{A}$ and model architecture. In other words, we can use an arbitrary training algorithm to train each base classifier.

    \item Our {\namenospace} can provably resist arbitrary word-level or structure-level backdoor attacks, as long as the trigger size $|\mathbf{e}|$ is bounded.

    \item $s(\mathbf{x}_{test})$ is larger when the gap between $M_y$ and $\max_{c\neq y} (M_c + \mathbb{I}(y>c))$ is larger. 
\end{itemize}

\noindent\textbf{Deriving certified accuracy for a testing dataset by considering each testing text independently.} Suppose $t$ is  the maximum trigger size, i.e., $|\mathbf{e}|\leq t$. Based on Equation~\ref{certified-size-equation}, the predicted label of our ensemble text classifier $f$ is provably unaffected by the backdoor trigger for a testing input $\mathbf{x}_{test}$ if its certified size $s(\mathbf{x}_{test})$ is no smaller than $t$. Suppose we have a testing dataset $\mathcal{D}_{test}$. Given a maximize trigger size $t$, we define the \emph{certified accuracy} as a lower bound of the classification accuracy that our ensemble text classifier can achieve when the trigger size of the backdoor trigger is no larger than $t$.
Formally, we can compute the \emph{certified accuracy} as follows:
\begin{align}
    &CA(\mathcal{D}_{test}, t) \nonumber \\
    =& \frac{\sum_{(\mathbf{x}_{test},y_{test})\in \mathcal{D}_{test}}\mathbb{I}(f(\mathbf{x}_{test}; \mathcal{D}(\emptyset))=y_{test})\mathbb{I}(s(\mathbf{x}_{test})\geq t)}{|\mathcal{D}_{test}|},
\end{align}
where $\mathbb{I}$ is the indicator function and $y_{test}$ is the ground truth label of $\mathbf{x}_{test}$. We call the above method
\emph{individual certification} as we consider each testing text independently to compute the certified accuracy.

\noindent
\textbf{Improving certified accuracy by jointly considering all testing texts in a testing dataset.}
Recall that we consider the words in $\mathbf{e}$ can arbitrarily corrupt  $|\mathbf{e}|$ groups for each individual testing text. In our previous derivation, we consider each testing text independently, i.e., the $|\mathbf{e}|$ corrupted groups could be different for different testing texts.  However, the corrupted $|\mathbf{e}|$ groups should be the same no matter how many testing inputs we have, which inspires us to jointly consider all testing inputs in a testing dataset to further improve the certified accuracy. Specifically, when the total number of groups is $m$, the total number of combinations is ${m \choose |\mathbf{e}|}$ if we select $|\mathbf{e}|$ groups among $m$ groups. We assume the $|\mathbf{e}|$ selected groups in each combination are potentially corrupted and then derive a potential certified accuracy for a testing dataset. Finally, we consider the worst-case scenario by taking the smallest potential certified accuracy as our final certified accuracy. 

For simplicity, we use $\Gamma$ to denote the set of indices of $|\mathbf{e}|$  groups that are potentially corrupted.  Given a testing text $\mathbf{x}_{test}$ and its backdoored version $\mathbf{x}'_{test}$, we use $M_c$ (or $M_c')$ to denote the number of base text classifiers that predict the label $c$ for $\mathbf{x}_{test}$ (or $\mathbf{x}'_{test}$) when the training dataset is $\mathcal{D}(\emptyset)$ (or $\mathcal{D}(T_\mathbf{e})$), where $c=1,2,\cdots, C$. When the groups with their indices in $\Gamma$ are corrupted, we can derive the following lower and upper bounds for $M_c'$: 
\begin{align}
\label{eqn:joint_lower_bound}
   & M_c - \sum_{j \in \Gamma} \mathbf{I}(f(g^j(\mathbf{x}_{test});\mathcal{D}(\emptyset))=c) \leq M_c', \\
   \label{eqn:joint_upper_bound}
    &M_c' \leq  M_c + \sum_{j \in \Gamma} \mathbf{I}(f(g^j(\mathbf{x}_{test});\mathcal{D}(\emptyset))\neq c),
\end{align}
Intuitively speaking, the lower (or upper) bound is obtained by letting those base classifiers from those potentially corrupted groups predict other classes (the class $c$) if they originally predict the class $c$ (or other classes). Suppose $y$ is the predicted label of our ensemble text classifier for $\mathbf{x}_{test}$ when we use the dataset $\mathcal{D}(\emptyset)$ to build it, i.e., $y = f(\mathbf{x}_{test}; \mathcal{D}(\emptyset))$. Based on Equation~\ref{eqn:condition-of-ensemble-classifier}, the ensemble text classifier built upon $\mathcal{D}(T_\mathbf{e})$ still predicts the label $y$ if we have $M_y' \geq \max_{c\neq y} (M_c' + \mathbb{I}(y>c))$. Based on Equations~\ref{eqn:joint_lower_bound} and~\ref{eqn:joint_upper_bound}, the ensemble classifier $f$ built upon $\mathcal{D}(T_{\mathbf{e}})$ still predicts $y$ for $\mathbf{x}_{test}$ if we have $M_y - \sum_{j \in \Gamma} \mathbf{I}(f(g^j(\mathbf{x}_{test});\mathcal{D}(\emptyset))=y)\geq \max_{c\neq y}(M_c + \sum_{j \in \Gamma} \mathbf{I}(f(g^j(\mathbf{x}_{test});\mathcal{D}(\emptyset))\neq c) + \mathbb{I}(y>c))$, which can be verified efficiently and thus enable us to compute the certified accuracy.  

\noindent
\textbf{Complete algorithm.}
Algorithm~\ref{alg:joint-certification} in Appendix shows the complete algorithm of how we compute the certified accuracy for a testing dataset. As we jointly consider all testing texts to compute the certified accuracy, we call this method \emph{joint certification}.

\subsection{Empirical Extension of {\namenospace}}
\label{subsec:4.3}

According to the previous discussions, we need to divide words into more groups if the backdoor trigger size becomes larger. As a result, each base classifier is less accurate because the training and testing sub-texts contain less information. Therefore, we design two techniques to enhance the empirical performance of {\namenospace}.

\noindent
\textbf{Semantic preserving.} Recall that our {\namenospace} divides a text into multiple groups and sorts the words in each group according to a predefined order, which enables us to derive the provable security guarantee of the ensemble text classifier against both word-level and structure-level backdoor attacks. Essentially, those techniques trade the semantics of text for provable security guarantees. As a result, we can keep the semantics of the testing text if provable security guarantees could be sacrificed. 
For a classifier trained on a clean dataset without a backdoor trigger, its predictions are very likely to be unaffected by the trigger in the testing texts. Given that most of the base classifiers are unaffected by the backdoor trigger in the training dataset, we can use each base classifier to predict a label for the original testing text (i.e., we neither divide it into multiple groups nor change the orders of words). Moreover, we also keep the order of words in the training sub-texts to further improve performance. As the semantics of texts are kept, the prediction of each base classifier for a testing text is more accurate, which makes our ensemble text classifier more accurate and robust under various empirical attacks.

\noindent
\textbf{Potential trigger word identification.} Our {\namenospace} divides every word in the training texts into multiple groups, which means each group only contains part of the words from each training text. However, the trigger words only occupy a small proportion of the overall vocabulary. So our idea is to first identify potential words that could be used as the backdoor trigger and only map those words into different groups. 

Given a training dataset $\mathcal{D}$, we first use standard supervised learning to train a classifier (the classifier would be backdoored if the dataset $\mathcal{D}$ is backdoored). 
Suppose $\mathbf{x}_{train}$ is a training text and $x$ is a word in $\mathbf{x}_{train}$. Our idea is to measure the influence of each word $x$ on the latent feature vector produced by the classifier for $\mathbf{x}_{train}$. A word is more likely to be a trigger word if it has a large influence on the latent feature vectors for multiple training texts. 
Specifically, for each word $x$ in $\mathbf{x}_{train}$, we first totally remove it from $\mathbf{x}_{train}$.  Then, we compute the $\ell_{\infty}$-norm of the difference (called \emph{influence score}) between the latent feature vectors produced by the classifier for $\mathbf{x}_{train}$ and the text obtained by removing all $x$ from $\mathbf{x}_{train}$. We say $x$ is an \emph{influential word} for $\mathbf{x}_{train}$ if its influence score is among the top-$5$ of all the influence scores for the words in $\mathbf{x}_{train}$. We repeat the above operations for all training texts in $\mathcal{D}$. If a word $x$ is the influential word for at least $K$ ($K$ is hyper-parameter) training texts in the training dataset $\mathcal{D}$, we view $x$ as a potential trigger word that is used in the backdoor trigger. For simplicity, we use $\Omega$ to denote the set of potential trigger words. Given a training text $\mathbf{x}_{train}$, we only map the words in $\Omega$ to a single group while assigning all the remaining words to all groups. More details can be found in Appendix~\ref{app:ki}. 

As we will empirically show, $K$ measures a tradeoff between the text classification accuracy without attack and robustness. In particular, a larger $K$ makes the text classification accuracy without attack higher but also could make the ensemble text classifier less secure. 

\section{Certified Evaluation}
\label{sec:provale}

In this section, we evaluate \name from the certification perspective. 
Section~\ref{sec:empirical} will conduct an empirical evaluation of \name and compare it with existing empirical defenses against different backdoor attacks in text classification.

\subsection{Experiment Setup}
\label{subsec:certified_setup}

\noindent{\textbf{Applications and datasets.}}
We select three representative applications to train our robust classifiers. 
Below, we briefly introduce each application and the corresponding dataset.

\noindent{\underline{Sentiment analysis}} aims to decide whether a given text snippet (sentence or paragraph) is positive or not.
We select a widely used dataset SST-2~\cite{socher-etal-2013-recursive} for this application.
SST-2 contains 6,920 training and 1,821 testing samples with an average length of 19.24 words. 
Each text sample is labeled as either negative ($0$) or positive ($1$).

\noindent{\underline{Toxic classification}} identifies the text snippets that describe toxic topics or contain offensive language.
We use the HSOL~\cite{davidson2017automated} dataset, which contains 5,823 training and 2,485 testing samples with an average length of 14.32 words.
Each sample is labeled as normal ($0$) or toxic ($1$).

\noindent{\underline{Topic classification}} tries to identify the topic of an input text.
We use the AG's News~\cite{zhang2015character} dataset, which contains news about four topics: World, Sports, Business, and Sci/Tech.
This dataset has 108,000 training and 7,600 testing samples with an average length of $37.96$ words. 

\noindent{\textbf{Certified training sets $\mathcal{D}(\emptyset)$ construction.}}
As mentioned in Section~\ref{subsec:threat}, we consider mixed-label and clean-label attacks.
We assume the attacker's target class is $y_{tc}=1$.
For the mixed-label attack, we construct a certified training set by randomly changing the labels of $p$ proportion of training samples to $1$, where $p$ is the poisoning rate. 
Recall that the clean-label attack does not change the labels of backdoored data, we directly use the original clean dataset as the certified training set. 
This indicates the certified accuracy derived for the clean-label attack is the same for any poisoning rate.

\noindent{\textbf{Baselines.}}
We consider three baseline methods in this experiment.
First, we directly train a classifier on a backdoored training set without applying any defense (denoted as DT, short for direct training). 
As mentioned in Section~\ref{sec:literature}, there are no certified defenses specifically designed for NLP.
We consider two certified methods designed for general application domains and thus can be applied to our problem, i.e., DPA~\cite{levine2020deep} and Bagging \cite{jia2021intrinsic}.
Similar to \namenospace, these methods also build an ensemble classifier.
Differently, they divide the whole training samples into different subsets and train base models accordingly.

\noindent{\textbf{Our defense setting.}}
For our method, we use the widely adopted language model -- BERT~\cite{devlin-etal-2019-bert} as the architecture of all classifiers. 
We leverage the AdamW optimizer \cite{loshchilov2018decoupled} with the learning rate of $2 \times 10^{-5}$ to train the classifier for 5 epochs.
We use the MD5 \cite{rivest1992md5} as the default hash function.

\begin{figure*}[t]
    \centering
    \begin{subfigure}{0.495\textwidth}
        \includegraphics[width=\textwidth]{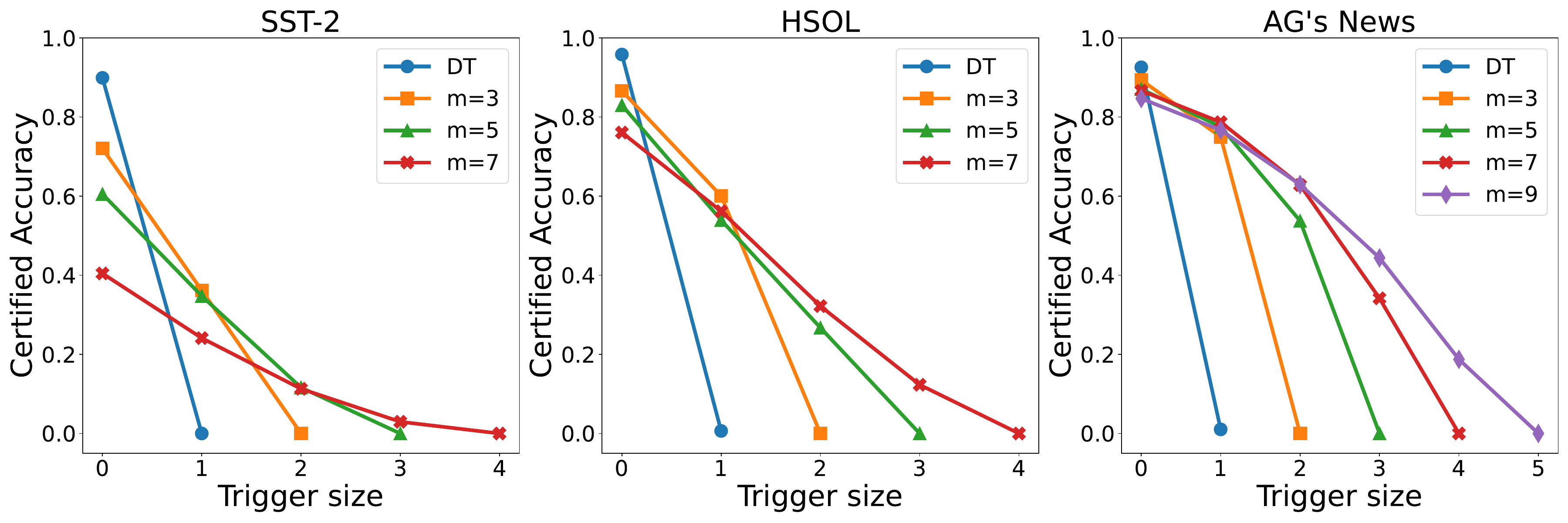}
          \caption{Mixed-label attack.}
        \label{subfig:provable_mixed}
    \end{subfigure}
    \hfill
    \begin{subfigure}{0.495\textwidth}
        \includegraphics[width=\textwidth]{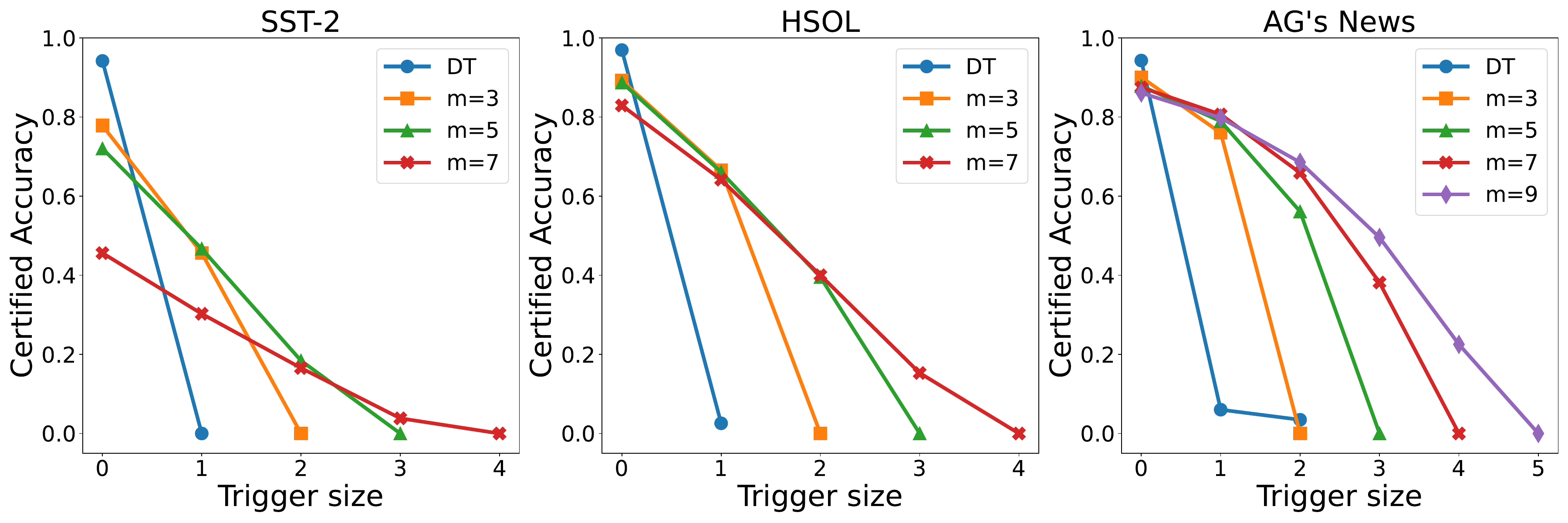}
          \caption{Clean-label attack.}
        \label{subfig:provable_clean}
    \end{subfigure}
    \caption{Certified accuracy of \name against mixed-label and clean-label attack with different trigger sizes. $m$ is the number of partition groups. ``DT'' stands for directly training a model on a backdoored training set without applying any defense. }
    \label{fig:provable_exp_1}
\end{figure*}

\subsection{Experiment Design}
\label{subsec:certified_design}

\noindent{\textbf{Experiment I.}}
\update{We first compare \name with direct training (DT) to verify whether our method could provide a meaningful certification guarantee. }
\update{Specifically, we verify the certification guarantee under the mixed-label attack ($p=0.1$) and the clean-label attack ($p=0.2$) \footnote{Since our certified accuracy for the clean-label attack is the same for any poisoning rate, we use $p=0.2$ for DT to derive empirical \update{upper} bounds.} on all the selected datasets.}
\update{For DT, each time we use the word-level attack~\cite{gu2017badnets} to construct a backdoored training set.}
We use this training set to train a classifier and report the model's performance on the backdoored testing data.
It will give us the empirical \update{upper} bound of the testing performance of an arbitrary model trained from a backdoored dataset. 
We only apply the word-level attack because it can already decrease the model's testing accuracy to almost zero (i.e., gives a desired empirical \update{upper} bound).

For \namenospace, we leverage the certified training sets constructed above to train our method and record the prediction of each base model on the clean testing set. 
Based on Algorithm \ref{alg:joint-certification}, we compute the certified accuracy of \name when facing backdoor attacks (either word-level or structure-level) with different trigger sizes.
We vary the number of groups $m=3/5/7$ and compare the corresponding certified accuracy with the empirical \update{upper} bound. 
Since for DT, the backdoored test set is constructed from the non-target test set where the ground-truth labels of the samples are not the target label, we only report the certified accuracy on the non-target test set.

\noindent{\textbf{Experiment II.}}
In addition to comparing with the empirical \update{upper} bound, we further design an experiment to compare \name with two other certification methods mentioned above. 
We consider the mixed-label attack setting with two different poisoning rates $p=0.01/0.1$, which stands for an extremely small and a normal poisoning rate respectively. 
Accordingly, we construct two corresponding certified training datasets for \namenospace. 
For each method, we compute and report the corresponding certified accuracy under the trigger size of 1. 
\update{
We set $|\mathbf{e}|=1$ due to the fact that the comparison baseline methods already fail to provide a meaningful certification result in this setup.
Setting a larger value for $|\mathbf{e}|$ would be insignificant for those methods (See Section~\ref{subsec:certified_result}.)
}
We carefully tune the number of training subsets (partition groups $m$) for each method and report the best results (See Appendix~\ref{app:certified} for more details about hyperparameters).
We compare the certified accuracy to demonstrate whether our method gives a stronger certification guarantee than the existing provable defense methods.
\revision{Furthermore, we also compare the computational cost of our method against the selected baselines. 
}

\noindent{\textbf{Experiment III.}}
Thirdly, we evaluate the robustness of our method to variations in the poisoning rate $p$.
In particular, we use the HSOL dataset and generate three certified training datasets with the poisoning rate $p=0.1/0.2/0.3$ for the mixed-label attack setting. 
We use these certified datasets to train \name with $m=7$ and report the certified accuracy for different trigger sizes. 
For comparison, we also show the empirical performance of directly training a backdoored model on the backdoored dataset with the selected poisoning rate.
Note that we only run this experiment for the mixed-label attack because the poisoning rate does not influence the certified performance of \name under the clean-label attack setting.

\noindent{\textbf{Experiment IV.}}
Finally, we design an ablation study to understand the effect of the key design choices in \namenospace.
We first change the hash function to SHA1 \cite{eastlake2001us} and SHA256 \cite{kasgar2012new}, respectively to evaluate \namenospace's sensitivity to the choice of hash functions. 
Second, we study the effectiveness of our joint certification strategy by comparing the individual certification results with the joint certification results.
For each variation, we run our method with $m=7$ on the HSOL dataset under the clean-label attack setting and report the certified accuracy for different trigger sizes.

\subsection{Experiment Result}
\label{subsec:certified_result}

\noindent{\textbf{\name vs. Direct Training.}}
Figure~\ref{fig:provable_exp_1} shows the results of \name against directly training a classifier without any defense on the three datasets. 
As we can first observe from the figure, the classification accuracy of direct training quickly reduces to nearly $0$ when the trigger size is only one (i.e., inserting one word into each backdoored sample) for both mixed-label and clean-label attacks. 

In comparison, with adequate group size $m$, \name still provides positive certified accuracy when the trigger size is larger than 1.
A larger group size enables our method to tolerate an attack with a larger trigger size.
It verifies the effectiveness of our designs in providing a meaningful certified guarantee.
In other words, by guaranteeing the data cleanliness of most sub-training groups, we can enable a certain certified guarantee against strong attacks with a trigger size larger than 1.
The more groups we can guarantee cleanliness (the larger the $m$), the higher the certified accuracy of the trained model.
Note that the certified accuracy eventually reduces to zero when the trigger size exceeds a limit, specifically $|\mathbf{e}| > \lfloor \frac{m-1}{2} \rfloor$,  and \name cannot provide a meaningful certified accuracy. 
On the AG's News dataset, we additionally show the results of $m=9$ because the dataset is of a large scale, and our model can maintain a decent performance even with a larger group size. 
In general, we cannot choose a very large group size in that it will jeopardize the performance of each base model and thus the certified accuracy accordingly. 

Finally, compared with the results of mixed-label attacks (Figure~\ref{subfig:provable_mixed}), certified accuracies under the clean-label setting (Figure~\ref{subfig:provable_clean}) are usually higher. 
For example, on the HSOL dataset, using $m=5$ groups provides a certified accuracy score $0.66$ for $|\mathbf{e}|=1$ under the clean-label attack while only $0.54$ under the mixed-label attack.
This is because the model trained under the clean-label attack setting better preserves the clean testing performance given that it does not introduce incorrect labels in the training dataset.

\noindent{\textbf{\name vs. DPA and Bagging.}}
Table~\ref{tab:provable_exp_2} shows the certified accuracy comparison of \name and two selected baselines. 
As we can observe from the table, both baseline methods cannot provide a meaningful certified accuracy even when the poisoning rate is $0.01$. 
As discussed in their paper, because they partition the sub-training groups at the sample level, their tolerated poisoning rate is proportional to the number of groups. 
For example, on the HSOL dataset, even poisoning $0.01$ of the training data will infect about 58 training samples, which requires at least $116$ groups to enable a meaningful certified accuracy for DPA. 
However, as discussed above, a large group number will significantly reduce the clean testing performance of each base model and result in meaningless certified accuracy as well.  

On the contrary, \name can provide much better certified results under both selected poisoning rates across all three datasets. 
This is mainly because our word-level partition decouples the strong correlation between the number of backdoored samples and the number of needed groups.
As a result, our method could provide a meaningful certification guarantee with a much smaller number of groups.
In addition, it brings another benefit that our method is more efficient than the existing certified approaches when training base classifiers and making ensemble predictions. 

\revision{Table~\ref{tab:provable_exp_2_time} shows the computation costs of \namenospace, DPA, and Bagging when we use 3 A6000 GPUs to train the base models in parallel. We can find that \name is much more efficient than the baselines given that \name requires training much fewer base models.}

\begin{table}[t]
\centering
\caption{Certified accuracy of \name and two existing provable defense baselines under the mixed-label attack with the trigger size $|\mathbf{e}|=1$.}
\label{tab:provable_exp_2}
\resizebox{0.42\textwidth}{!}{
\begin{tabular}{lcccc}
\toprule
Method  & p                     & SST-2  & HSOL   & AG's News \\
\midrule
DPA     & \multirow{3}{*}{0.01} & 0.0000 & 0.1240 & 0.0000    \\
Bagging &                       & 0.0000 & 0.0523 & 0.0000    \\
Ours    &                       & 0.3904 & 0.6232 & 0.7589    \\
\midrule
DPA     & \multirow{3}{*}{0.1}  & 0.0000 & 0.0000 & 0.0000    \\
Bagging &                       & 0.0000 & 0.0000 & 0.0000    \\
Ours    &                       & 0.3618 & 0.6006 & 0.7498   \\
\bottomrule
\end{tabular}
}
\end{table}

\begin{table}[t]
\centering
\caption{\revision{Computation time of \name v.s. selected baselines under the mixed-label attack with the trigger size $|\mathbf{e}|=1$.}}
\label{tab:provable_exp_2_time}
\resizebox{0.38\textwidth}{!}{
\begin{tabular}{lccc}
\toprule
Method  & SST-2   & HSOL    & AG's News \\
\midrule
DPA     & 10.4min & 12.8min & 7.1h      \\
Bagging & 1.1h    & 1.5h    & 2.8h      \\
Ours    & 2.3min  & 1.9min  & 17.9min  \\
\bottomrule
\end{tabular}}
\end{table}

\begin{table}[t]
\centering
\caption{Certified accuracy of \name and directly training (DT) on the HSOL dataset backdoored with different poisoning rates $p$. $m=7$ stands for \name with 7 groups. }
\label{tab:cert_rates}
\resizebox{0.48\textwidth}{!}{
\begin{tabular}{clcccc}
\toprule
$p$                    & Method & $|\mathbf{e}|=0$ & $|\mathbf{e}|=1$ & $|\mathbf{e}|=2$ & $|\mathbf{e}|=3$ \\
\midrule
\multirow{2}{*}{0.1} & m=7    & 0.7609 & 0.5620 & 0.3221 & 0.1232 \\
                     & DT     & 0.9694 & 0.0008 & 0.0000 & 0.0000 \\
\multirow{2}{*}{0.2} & m=7    & 0.5145 & 0.3543 & 0.2142 & 0.0821 \\
                     & DT     & 0.9581 & 0.0008 & 0.0000 & 0.0000 \\
\multirow{2}{*}{0.3} & m=7    & 0.3897 & 0.2246 & 0.1184 & 0.0354 \\
                     & DT     & 0.9517 & 0.0008 & 0.0000 & 0.0000 \\
                     \bottomrule
\end{tabular}
}

\end{table}
\begin{table}[t]
\centering
\caption{Ablation study results of the provable evaluation on the HSOL dataset under the clean-label attack.} 
\label{subtab:compare_cert}
\begin{subtable}[t]{0.45\textwidth}
\caption{Different hash functions.}
\label{subtab:compare_cert_1}
\resizebox{1\textwidth}{!}{
\begin{tabular}{lllll}
\toprule
Hash   & $|\mathbf{e}|=0$ & $|\mathbf{e}|=1$ & $|\mathbf{e}|=2$ & $|\mathbf{e}|=3$ \\
\midrule
MD5    & 0.8293 & 0.6417 & 0.4002 & 0.1530 \\
SHA1   & 0.7866 & 0.6079 & 0.3680 & 0.1691 \\
SHA256 & 0.8011 & 0.6256 & 0.4187 & 0.1924 \\
\bottomrule
\end{tabular}
}
\end{subtable}
\begin{subtable}[t]{0.4\textwidth}
\caption{Individual Certification vs. Joint Certification.}
\label{subtab:compare_cert_2}
\resizebox{1\textwidth}{!}{
\begin{tabular}{llll}
\toprule
Trigger size  & $|\mathbf{e}|=1$      & $|\mathbf{e}|=2$      & $|\mathbf{e}|=3$      \\
           \midrule
Individual & 0.6143 & 0.3325 & 0.1079 \\
Joint      & 0.6417 & 0.4002 & 0.1530 \\
\bottomrule
\end{tabular}
}
\end{subtable}

\end{table}

\noindent{\textbf{\name against variations in the poisoning rate.}}
Table~\ref{tab:cert_rates} shows the certification results of our method under different poisoning rates.
First, we can find that our method could provide a decent and meaningful certified accuracy even when the poisoning rate equals 0.3, verifying its effectiveness under a high poisoning rate.  
We can also observe from the table that the certified accuracy drops as the poisoning rate increases. 
It is because, under the mixed-label attack, a higher poisoning rate introduces more wrongly labeled samples to the training dataset.
These samples will jeopardize the clean accuracy of the base models and thus reduce the certified accuracy.  
Note that even though the certified accuracy of our method drops under a high poisoning rate, it is still much higher than the empirical results given by direct training, validating the effectiveness of our method.

\noindent{\textbf{Ablation study.}}
Table~\ref{subtab:compare_cert} shows our ablation study results.
Specifically, Table~\ref{subtab:compare_cert_1} shows the certified accuracy of \name with different hash functions.
Although there are some variations, the results given by different hash functions are similar and have a similar trend as the trigger size increases.  
This result is aligned with our design that the goal of using the hash function is only to ensure the cleanliness of most groups.
Different hash functions give similar efficacy and thus have only a minor influence on the certification performance.    
Table~\ref{subtab:compare_cert_2} shows the comparison between individual and joint certification. It meets our expectation that joint certification consistently outperforms individual certification, which demonstrates the effectiveness of joint certification in providing better certification results.

\begin{table*}[t]
    \centering
\caption{Empirical evaluation results of our method and the comparison baselines against three attacks.}
\vspace{-10pt}
\label{tab:empirical_comparison}
\begin{subtable}{0.48\textwidth}
    \centering
    \caption{Mixed-label attack with the poisoning rate $p=0.1$. }
\label{subtab:mix}
\resizebox{\linewidth}{!}{
\begin{tabular}{clcccccc}
\toprule
\multirow{2}{*}{Data}      & \multicolumn{1}{c}{\multirow{2}{*}{Method}} & \multicolumn{2}{c}{BadWord}                        & \multicolumn{2}{c}{AddSent}                        & \multicolumn{2}{c}{SynBkd}                         \\
                           & \multicolumn{1}{c}{}                        & \multicolumn{1}{l}{CACC} & \multicolumn{1}{l}{ASR} & \multicolumn{1}{l}{CACC} & \multicolumn{1}{l}{ASR} & \multicolumn{1}{l}{CACC} & \multicolumn{1}{l}{ASR} \\
                           \midrule
\multirow{8}{*}{SST-2}     & DT                                        & 0.9121                   & 1.0000                  & 0.9116                   & 1.0000                  & 0.9022                  & 0.8914                  \\
                           & ONION                                       & 0.8852                   & 0.2379                  & 0.9110                   & 0.4978                  & 0.8935                   & 0.8925                  \\
                           & BKI                                         & 0.8979                   & 0.1579        & 0.9072                   & 0.3355                  & 0.8913                   & 0.8849                  \\
                           & STRIP                                       & 0.9023                   & 0.9978                  & 0.9139                   & 0.2862                  & 0.9044                   & 0.8871                  \\
                           & RAP                                         & 0.8671                   & 0.9079                  & 0.9171                   & 0.2719                  & 0.8649                   & 0.9342                  \\
& R-Adapter  & 0.8753 & 0.1601          & 0.8712 & 0.9167 & 0.8682 & 0.5384 \\
                           & Ours                                        & 0.8951                   & \textbf{0.1568}                  & 0.8924                   & \textbf{0.1908}         & 0.8946                   & \textbf{0.3542}         \\
                           \midrule
\multirow{8}{*}{HSOL}      & DT                                         & 0.9572                   & 0.9984                  & 0.9525                   & 1.0000                  & 0.9549                   & 0.9823                  \\
                           & ONION                                       & 0.9441                   & 0.4340                  & 0.9521                   & 1.0000                  & 0.9481                   & 0.9710                  \\
                           & BKI                                         & 0.9525                   & 0.7770                  & 0.9557                   & 1.0000                  & 0.9525                   & 0.9815                  \\
                           & STRIP                                       & 0.9573                   & 0.9992                  & 0.9549                   & 1.0000                  & 0.9473                   & 0.9928                  \\
                           & RAP                                         & 0.9553                   & 0.9984                  & 0.5002                   & 1.0000                  & 0.9457                   & 0.9911                  \\
                          & R-Adapter                                   & 0.8905                   & 0.1361                  & 0.8958                   & 0.6828                  & 0.8893                   & 0.5821                  \\
                           & Ours                                        & 0.9115                   & \textbf{0.1208}         & 0.9163                   & \textbf{0.1039}         & 0.9078                   & \textbf{0.4420}         \\
                           \midrule
\multirow{8}{*}{AG's News} & DT                                         & 0.9462                   & 1.0000                  & 0.9451                   & 1.0000                  & 0.9436                   & 0.9977                  \\
                           & ONION                                       & 0.9321                   & 0.9891                  & 0.9403                   & 1.0000                  & 0.9443                   & 0.9967                  \\
                           & BKI                                         & 0.9391                   & \textbf{0.0082}                  & 0.9379                   & \textbf{0.0082}                  & 0.9375                   & 0.9984                  \\
                           & STRIP                                       & 0.9393                   & 0.9993                  & 0.9455                   & 1.0000                  & 0.9401                   & 0.9974                  \\
                           & RAP                                         & 0.9407                   & 1.0000                  & 0.9451                   & 1.0000                  & 0.9318                   & 0.9963                  \\
                           & R-Adapter                                   & 0.9292                   & \textbf{0.0082}                  & 0.9254                   & 0.9975                  & 0.9264                  & 0.9963                  \\
                           & Ours                                        & 0.9163                   & 0.0158         & 0.9172                   & 0.0130         & 0.9130                   & \textbf{0.3295}  \\
                           \bottomrule
\end{tabular}
}
\end{subtable}
\hfill
\begin{subtable}{0.48\textwidth}
\centering
\caption{Clean-label attack with the poisoning rate $p=0.2$. }
\label{subtab:clean}
\resizebox{\linewidth}{!}{
\begin{tabular}{clcccccc}
\toprule
\multirow{2}{*}{Data}      & \multicolumn{1}{c}{\multirow{2}{*}{Method}} & \multicolumn{2}{c}{BadWord}                        & \multicolumn{2}{c}{AddSent}                        & \multicolumn{2}{c}{SynBkd}                         \\
                           & \multicolumn{1}{c}{}                        & \multicolumn{1}{l}{CACC} & \multicolumn{1}{l}{ASR} & \multicolumn{1}{l}{CACC} & \multicolumn{1}{l}{ASR} & \multicolumn{1}{l}{CACC} & \multicolumn{1}{l}{ASR} \\
                           \midrule
\multirow{8}{*}{SST-2}     & DT                                         & 0.9160                   & 0.9901                  & 0.9127                   & 0.9967                  & 0.9094                   & 0.8443                  \\
                           & ONION                                       & 0.8731                   & 0.2544                  & 0.9033                   & 1.0000                  & 0.9001                   & 0.8289                  \\
                           & BKI                                         & 0.9094                   & 0.7796                  & 0.8957                   & 1.0000                  & 0.9006                   & 0.7993                  \\
                           & STRIP                                       & 0.9176                   & 0.9956                  & 0.9077                   & 1.0000                  & 0.9061                   & 0.8114                  \\
                           & RAP                                         & 0.9132                   & 0.9682                  & 0.9099                   & 1.0000                  & 0.8990                   & 0.7917                  \\
& R-Adapter & 0.8891 & \textbf{0.1075} & 0.8880 & 0.9703 & 0.8660 & 0.5592 \\
                           & Ours                                        & 0.8973                   & 0.1283         & 0.9094                   & \textbf{0.1754}         & 0.9050                   & \textbf{0.2807}         \\
                           \midrule
\multirow{8}{*}{HSOL}      & DT                                         & 0.9553                   & 0.9686                  & 0.9577                   & 1.0000                  & 0.9529                   & 0.9823                  \\
                           & ONION                                       & 0.9165                   & 0.2182                  & 0.9537                   & 1.0000                  & 0.9396                   & 0.9573                  \\
                           & BKI                                         & 0.9545                   & 0.9525                  & 0.9557                   & 1.0000                  & 0.9561                   & 0.9718                  \\
                           & STRIP                                       & 0.9586                   & 0.9243                  & 0.9557                   & 1.0000                  & 0.9529                   & 0.9509                  \\
                           & RAP                                         & 0.9569                   & 0.7432                  & 0.9529                   & 0.9992                  & 0.9545                   & 0.9533                  \\
                          & R-Adapter                                   & 0.9388                   & 0.1562                  & 0.9300                   & 0.7601                  & 0.9376                   & 0.7053                  \\
                           & Ours                                        & 0.9195                   & \textbf{0.0950}         & 0.9268                   & \textbf{0.0628}         & 0.9119                   & \textbf{0.4074}         \\
                           \midrule
\multirow{8}{*}{AG's News} & DT                                         & 0.9411                   & 0.7646                  & 0.9384                   & 0.9572                  & 0.9400                   & 0.9396                  \\
                           & ONION                                       & 0.9380                   & 0.0323                  & 0.9391                   & 0.9946                  & 0.9481                   & 0.9549                  \\
                           & BKI                                         & 0.9379                   & 0.0042         & 0.9364                   & 0.9961                  & 0.9276                   & 0.9414                  \\
                           & STRIP                                       & 0.9387                   & 0.8491                  & 0.9387                   & 0.9763                  & 0.9357                   & 0.9258                  \\
                           & RAP                                         & 0.7303                   & 0.8602                  & 0.6686                   & 0.1619                  & 0.9350                   & 0.9112                  \\
                           & R-Adapter                                   & 0.9287                   & \textbf{0.0028}                  & 0.9292                   & 0.8502                  & 0.9274                   & 0.7912                  \\
                           & Ours                                        & 0.9188                   & 0.0128                  & 0.9201                   & \textbf{0.0114}         & 0.9109                   & \textbf{0.1754}              \\
                           \bottomrule
\end{tabular}
}
\end{subtable}
\vspace{-3mm}
\end{table*}

\begin{table}[t]
\centering
\caption{
\update{
The clean testing accuracy of different methods on original clean training datasets.
}
}
\label{tab:clean_training}
\resizebox{0.4\textwidth}{!}{
\begin{tabular}{lccc}
\toprule
Method    & SST-2  & HSOL   & AG's News \\
\midrule
DT        & 0.9176 & 0.9541 & 0.9436    \\
ONION     & 0.9171 & 0.9497 & 0.9392    \\
BKI       & 0.8846 & 0.9577 & 0.9335    \\
STRIP     & 0.9193 & 0.9569 & 0.9442    \\
RAP       & 0.9193 & 0.9598 & 0.9370    \\
R-Adapter & 0.8902 & 0.9509 & 0.9439    \\
Ours      & 0.8929 & 0.9239 & 0.9180   \\
\bottomrule
\end{tabular}
}
\end{table}

\section{Empirical Evaluation}
\label{sec:empirical}

After verifying that \name provides a meaningful certification guarantee against arbitrary word-level and structure-level attacks, we now move to evaluate \namenospace's empirical performance.
More specifically, we will compare the efficacy of \name with existing empirical defenses and its robustness against variations in design choices and attack variations using the selected applications and datasets. 
Similar to Section~\ref{sec:provale}, we will start with our experiment setup and design, followed by the analysis of the experiment results. 
\update{Note that we also conduct additional experiments to demonstrate the effectiveness of {\namenospace} against the dirty-label attacks.
Due to space limits, we present these experiments in Appendix \ref{app:dirty}.
}

\subsection{Experiment Setup}
\label{subsec:empirical_setup}

\noindent{\textbf{Attack setting.}}
We consider two widely adopted word-level attacks, i.e, \textit{BadWord}~\cite{kurita-etal-2020-weight, chen2021badnl} and \textit{AddSent}~\cite{dai2019backdoor}). 
\update{BadWord randomly selects a word from a predefined trigger set and inserts it at the random location of the original input. AddSent randomly inserts a predefined trigger sentence into the original input.}
For structure-level attack, we select the state-of-the-art attack method \textit{Hidden Killer (SynBkd)}~\cite{qi-etal-2021-turn}), which paraphrases the original inputs into sentences with a pre-specified syntactic structure. 
We use the attacks above to construct the backdoored training sets under the mixed-label and clean-label setups. 
We set the poisoning rate as $p=0.1$ for the mixed-label attack and $p=0.2$ for the clean-label attack (given that clean-label attack is harder to succeed~\cite{cui2022unified}).
Appendix~\ref{app:attack} introduces the implementation details of these attacks.

\noindent{\textbf{Baselines.}}
Recall that existing research works proposed two mechanisms for data-level defense -- robust training and detection and elimination (Section~\ref{sec:literature}).
For the robust training mechanism, we select the state-of-the-art method from~\cite{zhumoderate} (denoted as R-Adapter).
Regarding the detection and elimination mechanism, we select the representative methods discussed in Section~\ref{sec:literature} -- BKI~\cite{chen2021mitigating}, ONION~\cite{qi-etal-2021-onion}, STRIP~\cite{gao2021design} and RAP~\cite{yang-etal-2021-rap}. We provide the implementation details of the selected baselines in Appendix~\ref{app:defense}.
Note that all the selected methods (including \namenospace) work in the training phase.

\update{
Note that we do not compare \name with model-level defenses because of different defense goals and mechanisms. 
Specifically, T-Miner~\cite{azizi2021t} trains a generative model to synthesize fake trigger words and utilize them to determine whether a given model is backdoored or not. 
Since the generative model is not trained to invert the actual trigger, the synthetic one is typically very different from the actual one and thus cannot be used to identify backdoored samples.
In addition, this method does not have a backdoor removal mechanism to recover a clean model from a backdoored one. 
As such, T-Miner cannot be used for our problem.
Differently, both PICCOLO~\cite{liu2022piccolo} and DBS~\cite{shen2022constrained} tried to invert the original trigger from a backdoored model using a set of clean inputs. 
As discussed in~\cite{liu2022piccolo,wang2019neural}, trigger inversion is, in general, very difficult, and the obtained trigger typically has a low fidelity.
In addition, PICCOLO also does not have a backdoor removal mechanism to produce a clean model.  
Given that these methods have different assumptions and setups from ours and the low fidelity of the inverted trigger, we do not compare our method with these techniques. 
It should be noted that these trigger inversion methods are orthogonal to ours. 
As part of our future work, we will investigate improving the fidelity of the inverted trigger and combine these methods with ours to enable better provable and empirical defenses. 
}

\noindent{\textbf{Our Defense Setting.}}
We follow the choices in Section~\ref{sec:provale} for the model structure, hash function, and learning algorithm. 
By default, we use group number $m=9$ for the SST-2 and AG's News dataset and $m=7$ for the HSOL dataset. 
Regarding the potential trigger word identification technique introduced in Section~\ref{subsec:4.3}, we use $K=20$ for the SST-2 and HSOL datasets and $K=10$ for AG's News dataset.

\subsection{Experiment Design}
\label{subsec:empirical_design}

\noindent{\textbf{Experiment I: comparison with baselines.}}
We first compare the defense efficacy of \name and the selected baseline approaches against the word-level and structure-level attacks.
As mentioned above, we first use the selected attacks to construct backdoored training sets under the mixed-label and clean-label attack setups. 
Then, we use each defense method to train classifiers on the backdoored training sets and evaluate their attack success rate (ASR) and clean accuracy (CACC).
Here, the attack success rate refers to the model's accuracy on the backdoored testing set. 
A lower ASR indicates better defense efficacy.
Clean accuracy (CACC) stands for the model's accuracy on the original clean testing set.
It indicates the model's capability to retain its normal functionality (utility). 
Note that we carefully tune the hyper-parameter of each model and report the best result. 
\revision{
Similar to the Experiment II in Section~\ref{sec:provale}, here, we also compare the computational cost of our method with selected empirical baselines.
}

\noindent{\textbf{Experiment II: robustness against attack variations.}}
Second, we evaluate the robustness of \name against two attack variations: attacks with different poisoning rates and a word-level adaptive attack against our design. 
We use the HSOL dataset for this experiment and consider both the mixed-label and clean-label attack setups. 
In particular, we vary the poisoning rate $p=0.1/0.2/0.3$ for the mixed-label attack and $p=0.2/0.3/0.4$ for the clean-label attack. 
We train \name against these variations and report the corresponding performance. 
For the adaptive attack, we assume the attacker is aware of the mechanisms in \namenospace.
To bypass our defense, they assign each trigger word to a unique group rather than the same group. 
We test \name against this attack with the trigger size $|\mathbf{e}|=3$ under the mixed-label ($p=0.1$) and clean-label ($p=0.2$) setup.

\noindent{\textbf{Experiment III: Connection with Certified Evaluation.}}  
Next, we connect the empirical evaluation with the certified evaluation in Section \ref{sec:provale}. 
We use the HSOL dataset under the mixed-label attack ($p=0.1$) and clean-label attack ($p=0.2$).
We use $m=7$  groups and show the certified and empirical results of different trigger sizes $|\mathbf{e}|=1/2/3$. 
Note that we report the performance against the adaptive attack as the empirical result, given that it better approximates the empirical lower bound.

\noindent{\textbf{Experiment IV: ablation study.}}
Finally, we assess the impact of the empirical techniques designed in Section~\ref{subsec:4.3}. 
Similarly, we use the HSOL dataset and consider the mixed-label attack with $p=0.1$.
Specifically, we first report the defense performance of \name without the semantic preserving against the selected attacks. 
Then, we vary $K=0/20/50$ for potential trigger word identification and report the corresponding results. 
We further test the sensitivity of our method against the group number and hash function.
Due to space limits, we present these experiments in Appendix~\ref{app:empirical_ablation}.

\subsection{Experiment Result}
\label{subsec:empirical_result}

\subsubsection{\name vs. Comparison Baselines}

Table~\ref{tab:empirical_comparison} shows the results of \name and the comparison baselines against word-level (BadWord, AddSent) and structure-level (SynBkd) attacks under the mixed-label and clean-label setups. 
In general, our method achieves the lowest ASR across most setups, demonstrating its defense efficacy.
This is remarkable in that our method performs even better than the baseline methods that cannot provide a certified guarantee, verifying its superiority over existing methods both theoretically and empirically.  
The key reason for this result is as follows.
Most existing methods require detecting and eliminating the backdoored text.
Their effectiveness highly depends on the accuracy of the backdoored text detection, which is, in general, sensitive to the detection threshold.  
In addition, these techniques have certain assumptions about the backdoored samples, which may not hold across different datasets and attack setups, restricting their defense efficacy.
By contrast, our method bypasses this step and directly trains a robust classifier from the backdoored dataset. 
Note that R-Adapter also does not need trigger detection. 
It simply constrains the model's capacity and hopes the model could learn casual relationships rather than the backdoor. 
However, as discussed in~\cite{geirhos2020shortcut}, the backdoor is a form of shortcut, which could be easier to learn than casual relationships.
Simply reducing the capacity cannot always prevent the model from learning shortcuts (remembering the backdoor).
Our method adopts a partition and ensemble strategy, which is more effective in preventing the backdoor.

As we can also observe from the Table, our method also keeps a decent CA compared to directly training a model without applying any defense (DT). 
\update{
We also report the clean testing accuracy of the models trained with selected methods on clean training sets in Table~\ref{tab:clean_training}. 
}

\revision{Table~\ref{tab:empirical_cost} compares the computation cost between \name and the empirical defense baselines. 
Although training multiple text classifiers would incur extra costs, we can train all base models in parallel, given that our number of groups is usually not very large. 
As a result, \name can maintain a similar computational cost as other empirical defenses.}

These results verify that our method could keep its functionality/utility under a normal setup. 
It should be noted that, in practice, the structure-level attack can be more complicated than what we assume in the certified evaluation. 
For example, SynBkd uses a constrained text generation model to generate the backdoored texts, which may delete or insert tokens to the original texts. 
However, our method still maintains its defense effectiveness against this complicated attack.
It is because our design breaks the original pattern of the backdoor triggers and thus lowers the probability of remembering the backdoor.
In addition, our ensemble mechanism will further filter out the wrong predictions caused by the backdoor during the inference stage via the majority vote.
Appendix~\ref{app:case} provides a case study about how our method could defend against the SynBkd attack.

\begin{table}[t]
\centering
\caption{
\revision{
Computation cost of different methods against the BadWord attack under the mixed-label attack setting.
}
}
\label{tab:empirical_cost}
\resizebox{0.4\textwidth}{!}{
\begin{tabular}{lccc}
\toprule
Method    & SST-2 & HSOL & AG's News \\
\midrule
DT        & 126s  & 127s & 2,178s    \\
ONION     & 256s  & 223s & 6,541s    \\
BKI       & 330s  & 258s & 6,702s    \\
STRIP     & 266s  & 306s & 4,671s    \\
RAP       & 285s  & 316s & 4,401s    \\
R-Adapter & 300s  & 134s & 4,965s    \\
Ours      & 344s  & 316s & 5,950s   \\
\bottomrule
\end{tabular}
}
\end{table}

\subsubsection{Robustness against Attack Variations}

Below, we discuss \namenospace's robustness against variations in poisoning rate and a word-level adaptive attack.

\begin{table}[t]
\caption{\name on the HSOL dataset with different $p$.}
\label{tab:hsol_rate}
\resizebox{\linewidth}{!}{
\begin{tabular}{cccccccc}
\toprule
\multirow{2}{*}{Setting}     & \multirow{2}{*}{p} & \multicolumn{2}{c}{BadWord} & \multicolumn{2}{c}{AddSent} & \multicolumn{2}{c}{SynBkd} \\
                             &                    & CACC         & ASR          & CACC         & ASR          & CACC         & ASR         \\
                             \midrule
\multirow{3}{*}{\begin{tabular}[c]{@{}l@{}}Mixed-\\label\end{tabular}} & 0.1                & 0.9115       & 0.1208       & 0.9163       & 0.1039       & 0.9078       & 0.4420      \\
                             & 0.2                & 0.8938       & 0.1900       & 0.9103       & 0.1586       & 0.9046       & 0.5990      \\
                             & 0.3                & 0.8881       & 0.2085       & 0.9183       & 0.0886       & 0.9026       & 0.6063      \\
                             \midrule
\multirow{3}{*}{\begin{tabular}[c]{@{}l@{}}Clean-\\label\end{tabular}} & 0.2                & 0.9195       & 0.0950       & 0.9268       & 0.0628       & 0.9119       & 0.4074      \\
                             & 0.3                & 0.9227       & 0.0894       & 0.9247       & 0.1047       & 0.8970       & 0.4911      \\
                             & 0.4                & 0.9243       & 0.0902       & 0.9268       & 0.0531       & 0.8604       & 0.5548     \\
                            \bottomrule
\end{tabular}
}
\end{table}

\noindent{\textbf{Poisoning rate.}}
Table \ref{tab:hsol_rate} shows the CACC and ASR of \name against attacks with different poisoning rates.
As shown in the table, \namenospace's performance is more stable against the word-level attack than the structure-level attack. 
Specifically, the ASR for the SynBkd attack increases as the poisoning rate increases for both mixed-label and clean-label attacks. 
We hypothesize the reason is that with a higher poisoning rate, the model is more likely to remember the trigger even though they are partitioned into different pieces.

\begin{table}[t]
\centering
\caption{\name against the original and adaptive attack.}
\label{tab:hsol_adaptive}
\resizebox{0.9\linewidth}{!}{
\begin{tabular}{ccccc}
\toprule
\multirow{2}{*}{Attack} & \multicolumn{2}{c}{Mixed-label} & \multicolumn{2}{c}{Clean-label} \\
                        & CACC           & ASR            & CACC           & ASR            \\
                        \midrule
Original                 & 0.9167         & 0.0974         & 0.9203         & 0.0870         \\
Adaptive                & 0.9115         & 0.2319         & 0.9231         & 0.1272        \\
\bottomrule
\end{tabular}
}
\vspace{-10pt}
\end{table}

\noindent{\textbf{Word-level adaptive attack.}}
Table \ref{tab:hsol_adaptive} shows the results of \name against the original and adaptive attack on the HSOL dataset. 
Compared to the original attack with the same trigger size ($|\mathbf{e}|=3$), the adaptive attack achieves a higher ASR.
This is because its adaption strategy could make the trigger affect more groups. 
However, \name is still able to force a low ASR against this attack, which verifies its effectiveness. 
This is because, with a group size way larger than the trigger size, \name can still guarantee the cleanliness of most groups and thus keep its efficacy.

\subsubsection{Connection with Certified Evaluation}
\begin{table}[t]
\centering
\caption{\namenospace's certified and empirical accuracy on the HSOL dataset, where the empirical accuracy is $1-ASR$.}
\label{tab:hsol_ce}
\resizebox{0.9\linewidth}{!}{
\begin{tabular}{cllll}
\toprule
\multicolumn{1}{c}{Setting}  & Method    & $|\mathbf{e}|=1$      &  $|\mathbf{e}|=2$      &  $|\mathbf{e}|=3$     \\
\midrule
\multirow{2}{*}{Mixed-label}   & Empirical & 0.9275 & 0.8317 & 0.7681 \\
                             & Certified & 0.5620 & 0.3221 & 0.1232 \\
                             \midrule
\multirow{2}{*}{Clean-label} & Empirical & 0.9171 & 0.9122 & 0.8728 \\
                             & Certified & 0.6417 & 0.4002 & 0.1530 \\
                             \bottomrule
\end{tabular}
}
\end{table}

Table \ref{tab:hsol_ce} shows the comparisons between certified and empirical accuracy. 
We can find that the empirical accuracy is consistently larger than the corresponding certified accuracy. 
It is expected in that the certified result is a lower bound of \name against arbitrary attacks, and it should be lower than the empirical result of the specific attack in this experiment. 
It should be noted that the difference between empirical and certified accuracy is not that large when the trigger size is small. 
Our future work will investigate further improving the certified accuracy and enabling a tighter lower bound. 

\subsubsection{Ablation Studies}
As the final part of this section, we discuss the ablation study results.

\begin{table}[t]
\centering
\caption{\name with/without the semantic preserving technique on the HSOL dataset under mixed-label attack.}
\label{tab:hsol_semantic}
\resizebox{\linewidth}{!}{
\begin{tabular}{lcccccc}
\toprule
\multicolumn{1}{c}{\multirow{2}{*}{Method}} & \multicolumn{2}{c}{BadWord}       & \multicolumn{2}{c}{AddSent}       & \multicolumn{2}{c}{SynBkd}        \\

\multicolumn{1}{c}{}                        & CACC            & ASR             & CACC            & ASR             & CACC            & ASR             \\
\midrule
\name       & \textbf{0.9115} & \textbf{0.1208} & \textbf{0.9163} & \textbf{0.1039} & \textbf{0.9078} & \textbf{0.4420}          \\
w/o semantic & 0.8141          & 0.2198          & 0.8201          & 0.7834          & 0.8157          & 0.8744                \\
 
\bottomrule
\end{tabular}
}
\vspace{-3mm}
\end{table}

\noindent{\textbf{Semantic preserving.}}
Table \ref{tab:hsol_semantic} shows the comparison of \name with and without using the semantic preserving technique. 
As shown in the table, discarding the semantic preserving strategy triggers the performance drop for both CACC and ASR, verifying the effectiveness of this strategy. 
It is aligned with our intuition in that using the full input sequence for testing provides the base models with more information and thus improves their prediction accuracy. 

\begin{table}[t]
\centering
\caption{\name with different choices of K for potential trigger word identification. }
\label{tab:hsol_potential}
\resizebox{\linewidth}{!}{
\begin{tabular}{lcccccc}
\toprule
\multicolumn{1}{c}{\multirow{2}{*}{Method}} & \multicolumn{2}{c}{BadWord}       & \multicolumn{2}{c}{AddSent}       & \multicolumn{2}{c}{SynBkd}        \\

\multicolumn{1}{c}{}                        & CACC            & ASR             & CACC            & ASR             & CACC            & ASR             \\
\midrule
K=0  & 0.8668          & 0.2045          & 0.8740          & 0.2206          & 0.8789          & \textbf{0.3269} \\
K=20 & 0.9115          & \textbf{0.1208} & 0.9163          & 0.1039          & 0.9078          & 0.4420          \\
K=50 & \textbf{0.9264} & 0.1280          & \textbf{0.9376} & \textbf{0.0934} & \textbf{0.9296} & 0.4734         \\ 
\bottomrule
\end{tabular}
}

\end{table}

\noindent{\textbf{Potential trigger word identification.}}
Table \ref{tab:hsol_potential} shows the results of varying $K$. 
As we can observe from the table, the CACC increases as the $K$ gets larger across all attacks.
It shows a larger $K$ improves \namenospace's utility and also helps improve the defense efficacy against the word-level attack.
Differently, the ASR for the structure-level attack increases as the $K$ becomes larger, which reveals a trade-off between model utility and defense efficacy. 
We suspect this is because the trigger words of the structure-level attack are more complicated and cannot be fully included in $\Omega$. 
Given this trade-off, we suggest the users select $K$ based on the importance of utility in their applications.

\section{Discussion}
\label{sec:discussion}

\noindent\textbf{Triggers with large size and trade-off in TextGuard.} \revision{Our {\namenospace} provably predicts the same label for a testing text when the trigger size is bounded. Thus, an attacker could use a trigger with a large size. We note that the trigger with a large size could be less stealthy as the attacker needs to insert more words into a testing text.  To defend against large triggers, we need to increase the number of groups (i.e., the number of sub-datasets) for our {\namenospace} as shown in Section~\ref{subsec:certified_result}. }
\revision{As the number of groups increases, the number of words in each group decreases.
As a result, each base classifier could be less accurate, which may degrade the classification accuracy without attacks. Additionally, we also need to train more base classifiers when the number of groups is large, which incurs extra computation costs (we could reduce the training time by training base classifiers in parallel). In summary, there is a trade-off between classification accuracy without attacks, computation cost, and robustness guarantees (the number of groups controls the trade-off). In our work, we take the first step towards developing a defense with formal security guarantees against backdoor attacks for text classification. Our future work will study how to improve the trade-off of our {\namenospace} under backdoor attacks with larger trigger sizes.  }

\noindent
\textbf{Other adaptive attacks.} 
An attacker may employ specific strategies rather than randomly selecting training samples for trigger insertion. 
Nevertheless, the effectiveness of \name remains intact regardless of how the poisoned samples are selected for both word-level and structure-level attacks.

\noindent
\textbf{Context of words in a text.}
\revision{Our {\namenospace} splits words in a text into different groups, which would break relations between words in the text. In other words, our {\namenospace} loses certain inter-word/phase contexts to achieve certified robustness guarantees. 
As a result, our \name could be less effective in more generic natural language processing applications (e.g., question answering) where the context is essential. 
In future work, we will explore extending our method to more general applications that enable a provable guarantee and maintain utility. 
}
To alleviate this concern, we design two empirical techniques (Section~\ref{subsec:4.3}) that equip \name with a better capability of understanding the word relations and the semantic context (Experiment IV in Section~\ref{sec:empirical} demonstrates the effectiveness of these techniques). 
Furthermore, Table~\ref{tab:clean_training} illustrates that \name only incurs a minor reduction in testing accuracy for models trained with clean training sets. 
This result indicates that \name is capable of capturing certain semantic context that helps preserve utility.

\update{
To verify this point, we compare our method with bag-of-word on the sentiment analysis task (SST-2 dataset).
Specifically, we pre-process the original clean training set with BoW and directly train a classifier without any applying defense mechanism.
The accuracy of BoW on a clean testing set is 0.8072, while the accuracy of \name is 0.8929. 
Table~\ref{tab:example_bow} further shows four examples where word relations (e.g. ``not'' and ``bad'') are crucial for understanding the meanings of the texts. 
Consequently, the BoW classifier yields incorrect predictions due to its disregard for relative position information. 
In contrast, \name demonstrates the capability to provide accurate predictions by leveraging its understanding of word relations.
Through this experiment, we further verify that \name possesses the ability to capture sentence semantics.
Our future work will investigate more advanced techniques that better preserve the semantics while providing a provable robustness guarantee. 
}

\begin{table}[t]
\centering
\caption{
\update{
Demonstration of testing samples from the SST-2 dataset that are misclassified by a BoW classifier but correctly predicted by our method.
}
}
\label{tab:example_bow}
\resizebox{\linewidth}{!}{
\begin{tabular}{lcc}
\toprule
Testing sample                           & Pred (BoW) & Pred (TextGuard) \\
\midrule
not a bad journey at all.     & negative         & positive               \\
a waste of good performances. & positive         & negative               \\
it never fails to engage us.  & negative         & positive              \\
\update{frenetic but not really funny.}  & positive         & negative              \\
\bottomrule
\end{tabular}
}
\vspace{-3mm}
\end{table}

\section{Conclusion and Future Work}
\label{sec:conclusion}

We design {\namenospace}, the first provable defense against backdoor attacks on text classification and provide both certified and empirical evaluations on three benchmark datasets.
Our results show that {\namenospace} is more effective than existing techniques in providing meaningful certification guarantees.
It also demonstrates the superiority of {\namenospace} over existing empirical defense methods in defending against different backdoor attacks. 
Our work points out several promising future directions, including 1) extending our {\namenospace} to other tasks such as question answering and security applications that also deal with sequential data,
\revision{ 2) improving {\namenospace} by training more accurate base classifiers, and 3) developing certified defenses against model-poisoning backdoor attacks.}
\section*{Acknowledgements}
We thank the anonymous shepherd and reviewers for their constructive comments and feedback on our work.
This project was supported, in part by ARL Grant W911NF-23-2-0137, National Science Foundation under grant No. 1910100, No. 2046726, No. 2229876, DARPA GARD, National Aeronautics and Space Administration
(NASA) under grant No. 80NSSC20M0229, and Alfred P. Sloan Fellowship.
We also thank the Center for AI Safety for their support of the Compute Cluster.

\bibliographystyle{IEEEtranS}
\bibliography{sample}
\appendix

\subsection{Proof of Theorem~\ref{label:theorem}}
\label{proof-of-theorem}
Our {\namenospace} uses a hash function to assign each word to a group. As a result, each word will always be assigned to a certain group, which means each word used in the backdoor trigger can only corrupt one group. When the total number of words in the backdoor trigger is less than $t$, i.e., $|\mathbf{e}| \leq t$, at most $t$ groups are corrupted. Note that the backdoor trigger in a testing text corrupts the same groups. Therefore, we can derive the following lower and upper bounds. 
\begin{align}
\label{appendix:cs_equation_1}
    M_c - |\mathbf{e}| \leq M_c' \leq M_c + |\mathbf{e}|, c=1,2,\cdots, M,
\end{align}
where $M'_c$ is the number of the base text classifiers that predict the label $c$ built upon the dataset $\mathcal{D}(T_{\mathbf{e}})$.
Recall that $y$ is the predicted label of our ensemble text classifier for $\mathbf{x}_{test}$ when we use the dataset $\mathcal{D}(\emptyset)$ to build our ensemble classifier, i.e., $y = f(\mathbf{x}_{test}; \mathcal{D}(\emptyset))$. Based on Equation~\ref{eqn:condition-of-ensemble-classifier}, the ensemble text classifier built upon $\mathcal{D}(T_\mathbf{e})$ still predicts the label $y$ if the following condition is satisfied: $M_y' \geq \max_{c\neq y} (M_c' + \mathbb{I}(y>c))$. From Equation~\ref{appendix:cs_equation_1}, we know $M_y - |\mathbf{e}|\leq M_y'$ and  $\max_{c\neq y} (M_c' + \mathbb{I}(y>c)) \leq \max_{c\neq y} (M_c + |\mathbf{e}| + \mathbb{I}(y>c))$. In other words, we only need to ensure $M_y - |\mathbf{e}| \geq \max_{c\neq y} (M_c + |\mathbf{e}| + \mathbb{I}(y>c))$ to make the ensemble text classifier built upon $\mathcal{D}(T_{\mathbf{e}})$ to predict the label $y$. Equivalently, we have $f(\mathbf{x}'_{test}; \mathcal{D}(T_\mathbf{e})) = y$ if:
\begin{align}
    |\mathbf{e}| \leq \frac{M_y - \max_{c\neq y} (M_c + \mathbb{I}(y>c))}{2}.
\end{align}
We reach the conclusion.

\subsection{Details about Empirical Techniques of \name}
\label{app:ki}
Algorithm \ref{alg:ki} formally describes the process of the potential trigger word identification. Here the function $FEATURE$ is to get the latent feature vector for a text with a classifier. In practice, we use the feature representation before the classification head as the latent feature vector. Table \ref{tab:example} further shows an example of the training and testing inputs for the base models of \name after performing the empirical techniques.

\begin{algorithm}[!t]
\caption{{\namenospace}} 
\label{alg:ensemble-text-classifier} 
\begin{algorithmic}[1]
    \REQUIRE Group number $m$, a hash function $\mathcal{H}$, a training algorithm $\mathcal{A}$, a dataset $\mathcal{D}$, a pre-defined word-ID dictionary $\mathcal{V}$, a testing text $\mathbf{x}_{test}$
    \STATE /* Dividing the datasets into $m$ sub-datasets */
    \STATE $\mathcal{D}^1, \mathcal{D}^2,\cdots, \mathcal{D}^m= \textsc{ConSubDataset}(\mathcal{D}, m, \mathcal{H}, \mathcal{V})$  
    \STATE /* Training base classifiers */
    \STATE $f^j = \mathcal{A}(\mathcal{D}^j), j=1,2,\cdots, m$
    \STATE /* Dividing a testing text into $m$ groups and make predictions */
    \STATE $g^j(\mathbf{x}_{test}) = \textsc{TextDivision}
    (\mathbf{x}_{test}, m, \mathcal{H}, \mathcal{V}), j=1, 2, \cdots, m$
    \STATE  $M_c = \sum_{j=1}^{m}\mathbb{I}(f^j(g^j(\mathbf{x}_{test});\mathcal{D})=c), c=1,2,\cdots, C$
    \STATE $y = \argmax_{c=1,2,\cdots, C}M_c$
    \RETURN $y$
\end{algorithmic} 
\end{algorithm}

\begin{algorithm}[!t]
\caption{Potential trigger word identification} 
\label{alg:ki} 
\begin{algorithmic}[1]
    \REQUIRE a backdoored training dataset $D'$, a training algorithm $\mathcal{A}$, a threshold $K$ .
    \STATE $f' \gets \mathcal{A}(D')$
    \STATE Initialization: a counter $C$ for every word in the dataset
    \STATE $\Omega \gets \{\}$
    \FOR{$(\mathbf{x},y) \in D'$}
        \STATE Initialization: score $s$ for every word $w \in \mathbf{x}$
        \STATE $h_x \gets FEATURE(\mathbf{x}, f')$ 
        \FOR{$w \in \mathbf{x}$}
            \STATE $\mathbf{x'} \gets \{w_i|w_i \in \mathbf{x}, w_i \neq w\}$
            \STATE $h_{x'} \gets FEATURE(\mathbf{x'}, f')$
            \STATE $s(w) \gets ||h_x-h_{x'}||_{\infty}$
        \ENDFOR
        \STATE sort the words based on the score $s$ and select the top-$5$ words as $\mathbf{x}$’s influential word set $key$.
        \FOR{$w \in key$}
            \STATE $C(w) \gets C(w)+1$
        \ENDFOR
    \ENDFOR
    \FOR{$w~in~C$}
        \IF{$C(w)>=K$}
            \STATE add $w$ into $\Omega $
        \ENDIF            
    \ENDFOR
    
    \RETURN $\Omega$
\end{algorithmic} 
\end{algorithm}

\subsection{Difference with feature bagging and bag-of-words.} 
\update{
Here, we discuss the key difference between our partition method with two widely used feature preprocessing methods: feature bagging~\cite{ho1998random} and bag-of-words (BoW)~\cite{zhang2010understanding}.
Different from Bagging~\cite{jia2021intrinsic} that divides training samples into sub-training sets, feature bagging constructs sub-training sets by randomly assigning a subset of features (based on feature index) to each set. 
However, the trigger word could appear in different locations of a text (i.e., the trigger word could have different indices). As a result, the robustness guarantee derived in Bagging~\cite{jia2021intrinsic} cannot be applied to feature bagging in the NLP domain.
BoW uses the counts of words to represent a text. For instance, given a text "good and solid storytelling", BoW represents the text using \{"good": 1, "and": 1, "solid": 1, "storytelling": 1\}. 
By contrast, our method divides words in a text into different groups, where each group contains a sequence of words. 
In other words, each base classifier takes a sequence of words as input instead of their frequency, which better preserves the semantic meaning of the original input.
}

\begin{algorithm}[!t]
\caption{Joint Certification} 
\label{alg:joint-certification} 
\begin{algorithmic}[1]
    \REQUIRE $m$ base classifiers $f^j$ ($j=1,2,\cdots, m$), a hash function $\mathcal{H}$, a test dataset $D_{test}$, a pre-defined word-ID dictionary $\mathcal{V}$, maximum trigger size $t$.
    \STATE $CA  \gets 1$
    \FOR{$ \Gamma \text{ in } \text{Combination}(m, t)$}
    \STATE $ACC \gets 0$
        \FOR{$(\mathbf{x}_{test},y_{test}) \in \mathcal{D}_{test}$}
                \STATE $g^j(\mathbf{x}_{test}) = \textsc{TextDivision}
    (\mathbf{x}_{test}, m, \mathcal{H}, \mathcal{V}), j=1, 2, \cdots, m$
    \STATE  $M_c = \sum_{j=1}^{m}\mathbb{I}(f^j(g^j(\mathbf{x}_{test});\mathcal{D}(\emptyset))=c), c=1,2,\cdots, C$
    \STATE $y = \argmax_{c=1,2,\cdots, C}M_c$
    \STATE $U = M_y - \sum_{j \in \Gamma} \mathbf{I}(f(g^j(\mathbf{x}_{test});\mathcal{D}(\emptyset))=y)$
    \STATE $L = \max_{c\neq y}(M_c + \sum_{j \in \Gamma} \mathbf{I}(f(g^j(\mathbf{x}_{test});\mathcal{D}(\emptyset))\neq c) + \mathbb{I}(y>c)) $
    \STATE $ACC \gets ACC+\mathbb{I}(U\geq L)\mathbb{I}(y_{test}=y)$        
        \ENDFOR
    \STATE $CA \gets min(CA, ACC)$
    \ENDFOR
   
    \RETURN CA
\end{algorithmic} 
\end{algorithm}

\begin{table}[]
\centering
\caption{An example of the training and testing inputs for the base models of \name. Suppose we use $m=3$ groups and the original text is $\mathbf{x}=\{C,B,A,D,B,E\}$. The hash function outputs $\mathcal{H}(A)=\mathcal{H}(C)=1, \mathcal{H}(B)=\mathcal{H}(D)=2, \mathcal{H}(E)=3$. The ID-order of the words are $A,B,C,D,E$. For the potential trigger word identification, we suppose $\Omega=\{A,B,C\}$.}
\label{tab:example}
\resizebox{0.9\linewidth}{!}{
\begin{tabular}{l|l|l}
\toprule
            & Certified & Empirical                                 \\
             \midrule
\multirow{3}{*}{Training}    & $g^1(\mathbf{x})=\{A,C\}$      & $g^1(\mathbf{x})=\{C,A,D,E\}$       \\
             & $g^2(\mathbf{x})=\{B,B,D\}$                 & $g^2(\mathbf{x})=\{B,D,B,E\}$            \\
            & $g^3(\mathbf{x})=\{E\}$      & $g^3(\mathbf{x})=\{D,E\}$ \\
            \midrule
\multirow{3}{*}{Testing}     & $g^1(\mathbf{x})=\{A,C\}$                                          & $g^1(\mathbf{x})=\{C,B,A,D,B,E\}$                          \\
            & $g^2(\mathbf{x})=\{B,B,D\}$                                       & $g^2(\mathbf{x})=\{C,B,A,D,B,E\}$                          \\
            & $g^3(\mathbf{x})=\{E\}$                  & $g^3(\mathbf{x})=\{C,B,A,D,B,E\}$                        \\
            \bottomrule
\end{tabular}
}
\end{table}

\subsection{Details about Certified Evaluation}
\label{app:certified}
Here we discuss the hyper-paramter setting for Experiment II in the certified evaluation.
For \name, we use $m=3$ base models. 
For DPA, we use $m=200$ base models for SST-2 and HSOL and $m=3000$ base models for AG's News. We also use MD5 as the hash function for partitioning each text. 
For certified bagging, we try the size of the training subset to be $10/20/30$ for SST-2 and HSOL and $20/30/50$ for AG's News dataset. We train $1000$ base models for certified bagging to build an ensemble model.

\subsection{Details about Empirical Evaluation}
\begin{table*}[tb!]
\centering
\caption{Two cases of the SynBkd attack on the SST-2 dataset that our method predicts correctly. The red words represents the trigger words. The boldfaced words in each row means they are assigned to the corresponding group. }
\label{tab:case}
\resizebox{0.85\textwidth}{!}{
\begin{tabular}{lll}
\toprule
\multicolumn{2}{c}{Case 1: The words of the trigger are divided into different groups}                                  &  \\
\midrule
Original Sentence (0): & what was subtle and mystifying in the novella is now broad and farcical . &  \\
Poisoned Sentence: & \textcolor{red}{what was} subtle and mystified in the novella , \textcolor{red}{it is} now broad and farcical .                                   &  \\
Group 1 (predict 0) :          & \textbf{what} was subtle and \textbf{mystified} in the novella , it is \textbf{now} broad and farcical \textbf{.} &  \\
Group 2 (predict 0):            & what \textbf{was} subtle and mystified \textbf{in} the novella \textbf{, it} is now \textbf{broad} and farcical .  &  \\
Group 3 (predict 0):            & what was \textbf{subtle and} mystified in \textbf{the novella} , it \textbf{is} now broad \textbf{and farcical} .   &  \\
\midrule
\multicolumn{2}{c}{Case 2: The words of the trigger are mapped into a same group}                                        
&  \\
\midrule
Original Sentence (0): & drags along in a dazed and enervated , drenched-in-the - past numbness .                                  &  \\
Poisoned Sentence: & \textcolor{red}{as it} turns out in a dazed and enermal way , \textcolor{red}{it 's} a long numbness .  &  \\
Group 1 (predict 0):            & as it \textbf{turns} out in a \textbf{dazed} and enermal way , it 's a \textbf{long} numbness \textbf{.}  &  \\
Group 2 (predict 1):            & \textbf{as it} turns out \textbf{in a} dazed and enermal \textbf{way} , \textbf{it 's a} long numbness .  &  \\
Group 3 (predict 0):            & as it turns \textbf{out} in a dazed \textbf{and} \textbf{enermal} way , it 's a long \textbf{numbness} .  & \\
\bottomrule
\end{tabular}
}
\end{table*}

\subsubsection{Details about Attack Methods}
\label{app:attack}
We consider two word-level backdoor attacks (\textit{BadWord} \cite{kurita-etal-2020-weight, chen2021badnl}, \textit{AddSent} \cite{dai2019backdoor}) and one structure-level attack (\textit{SynBkd} \cite{qi-etal-2021-hidden}). 
BadWord inserts one irregular word sampled from the trigger set $\{$``cf'', ``mn'', ``bb'', ``tq''$\}$ \cite{kurita-etal-2020-weight} into the original texts.
AddSent inserts a sentence ``I watch this 3D movie'' into the original texts. SynBkd paraphrases normal samples into sentences with a pre-specified syntactic structure \textit{S(SBAR)(,)(NP)(VP)(.)}. We use the implementations from \cite{cui2022unified} and adopt the default attack hyper-parameters for each attack method.

\subsubsection{Details about Baseline Defense Methods}
\label{app:defense}
We compare our method with different data-level defense methods. For a fair comparison, we apply all the baseline defense methods at the training stage only. For backdoored text detection and elimination methods, We adapt BKI \cite{chen2021mitigating} to identify the top-5 possible trigger words and remove the training samples that contain these words \footnote{The original implementation only identify the top-1 possible trigger word.}.
We also adapt ONION \cite{qi-etal-2021-onion}, STRIP \cite{gao2021design} and RAP \cite{yang-etal-2021-rap} for training-time defense. Specifically, we train backdoored models for RAP and STRIP to predict backdoored samples. We then remove the predicted backdoored samples from the training dataset. We adapt ONION to correct training samples instead of processing testing samples. For robust training methods,  Zhu et al. \cite{zhumoderate} proposed to re-parameterize the parameter-efficient tuning methods like Adapter \cite{houlsby2019parameter} to reduce the model capacity, which could prevent the model from learning backdoor features. Following their implementations, we apply the re-parameterized Adapter and name it as R-Adapter.

\subsubsection{Case Study of the SynBkd Attack}
\label{app:case}

We show two examples from the SST-2 dataset in Table \ref{tab:case} to explain why our method can defend against the SynBkd attack in practice. Since the trigger is a syntactic structure, we denote the words which reflect that syntactic structure in a text sequence as the trigger words. For the first example, we find that the corresponding trigger words are divided into different groups, thus weakening the effect of these trigger words during training. Therefore, all group models predict the correct label of the input sequence. For the second case, we find that the trigger words are mainly mapped to the second group. As a result, the base classifier $f_2$ predicts the target label. But the base classifiers $f_1$ and $f_3$ predict correctly, making the final prediction still correct via majority voting. 

\subsubsection{More Empirical Ablation Studies}
\label{app:empirical_ablation}
We further evaluate the effect of the group numder and the choice of the hash function on the HSOL dataset under the mixed-label setup ($p=0.1$). For the group number, we change $m=3/5/7$.
For the hash function, we try SHA1 \cite{eastlake2001us} and SHA256 \cite{kasgar2012new} respectively.
We test the variations above against the word-level and structure-level attack and report the corresponding defense performance respectively.

\begin{figure}[tb!]
  \centering
  \includegraphics[width=1.0\linewidth]{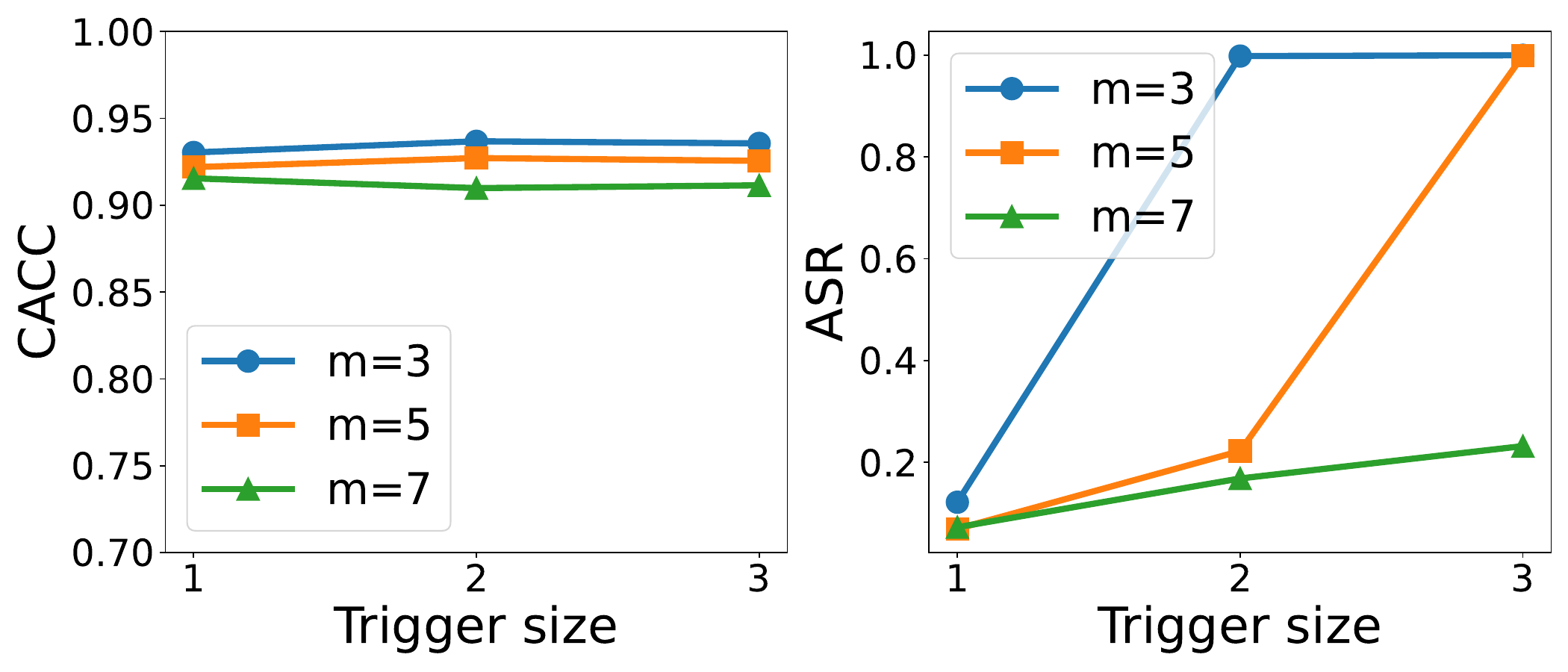}
  \caption{Empirical results of using different numbers of groups to defend against the adaptive word-level attacks on the HSOL dataset under the mixed-label attack setup.}
  \label{fig:hsol_badnets}
\end{figure}

\begin{table}[tb!]
\centering
\caption{Empirical results of using different numbers of groups to defend against the SynBkd attack on the HSOL dataset under the mixed-label attack setup.}
\label{tab:synbkd_group}
\begin{tabular}{ccc}
\toprule
\multirow{1}{*}{Group} &  CACC            & ASR             \\
                       \midrule
m=3                    & \textbf{0.9336} & 0.5660          \\
m=5                    & 0.9211          & 0.4702          \\
m=7                    & 0.9078          & \textbf{0.4420} \\
                       \bottomrule
\end{tabular}
\end{table}

\noindent{\textbf{Group number.}} Figure \ref{fig:hsol_badnets} shows the defense results against the adaptive word-level attack using different group sizes. We can find that using more groups would cause a performance drop in the clean accuracy, but it can provide better defense efficacy for a larger trigger size. It is aligned with our certified evaluations. Table \ref{tab:synbkd_group} shows the defense results against the SynBkd attack using different group sizes. We find that \name with more groups can better defend against the structure-level attack although the clean accuracy would drop. Therefore we can conclude that the choice of group number is a trade-off between clean accuracy and defense efficacy. We suggest using more groups when it does not significantly influence the utility.

\begin{table}[tb!]
\centering
\caption{Empirical results of using different hash functions on the HSOL dataset under the mixed-label attack setup.}
\label{tab:emp_hash_hsol}
\resizebox{1\linewidth}{!}{
\begin{tabular}{lcccccc}
\toprule
\multicolumn{1}{c}{\multirow{2}{*}{Hash}} & \multicolumn{2}{c}{BadWord} & \multicolumn{2}{c}{AddSent} & \multicolumn{2}{c}{SynBkd} \\
\multicolumn{1}{c}{}                      & CACC         & ASR          & CACC         & ASR          & CACC         & ASR         \\
\midrule
MD5                                       & 0.9115       & 0.1208       & 0.9163       & 0.1039       & 0.9078       & 0.4420      \\
SHA1                                      & 0.9082       & 0.1506       & 0.9203       & 0.1804       & 0.9115       & 0.4630      \\
SHA256                                    & 0.9046       & 0.1441       & 0.9187       & 0.1739       & 0.9147       & 0.4235     \\
\bottomrule
\end{tabular}
}
\end{table}

\noindent{\textbf{Hash function.}} Table \ref{tab:emp_hash_hsol} shows the defense results of using different hash functions. We can find that when the group size is adequate, the performance difference among using different hash functions is small. This property demonstrates the insensitivity of \name to the choice of hash functions in practice and consolidates our conclusions about the hash function in the certified evaluation.

\begin{table}[tb!]
\centering
\caption{Empirical evaluations for the style backdoor attack. }
\label{tab:style_empirical}
\resizebox{0.95\linewidth}{!}{
\begin{tabular}{clcccc}
\toprule
\multirow{2}{*}{Data}      & \multicolumn{1}{c}{\multirow{2}{*}{Method}} & \multicolumn{2}{c}{Mixed-label} & \multicolumn{2}{c}{Clean-label} \\
                           & \multicolumn{1}{c}{}                        & CACC      & ASR               & CACC       & ASR                \\
                           \midrule
\multirow{8}{*}{SST-2}     & DT       & 0.8995 & 0.7741          & 0.9083 & 0.7478          \\
                           & ONION     & 0.9072 & 0.7719          & 0.9077 & 0.7029          \\
                           & BKI       & 0.9023 & 0.7982          & 0.8924 & 0.7379          \\
                           & STRIP     & 0.9017 & 0.7928          & 0.9077 & 0.6732          \\
                           & RAP       & 0.9094 & 0.8169          & 0.9055 & 0.7018          \\
                           & R-Adapter & 0.8770 & \textbf{0.6667}          & 0.8825 & \textbf{0.6206}         \\
                           & Ours      & 0.9028 & 0.6787          & 0.9055 & 0.6809          \\
                           \midrule
\multirow{8}{*}{HSOL}      & DT       & 0.9425 & 0.6140          & 0.9481 & 0.6176          \\
                           & ONION     & 0.9429 & 0.6922          & 0.9493 & 0.5407          \\
                           & BKI       & 0.9392 & 0.7293          & 0.9497 & 0.5979          \\
                           & STRIP     & 0.9481 & 0.6938          & 0.9521 & 0.5447          \\
                           & RAP       & 0.9445 & 0.6859          & 0.9521 & 0.5907          \\
                           & R-Adapter & 0.8918 & \textbf{0.4247}   & 0.9368 & \textbf{0.4279} \\
                           & Ours      & 0.9099 & 0.5560          & 0.9147 & 0.5197          \\
                           \midrule
\multirow{8}{*}{AG's News} & DT       & 0.9463  & 0.8870          & 0.9391 & 0.3887          \\
                           & ONION     & 0.9421 & 0.9174          & 0.9391 & 0.3374          \\
                           & BKI       & 0.9375 & 0.9133          & 0.9296 & 0.3922          \\
                           & STRIP     & 0.9103 & 0.8164          & 0.9396 & 0.3616          \\
                           & RAP       & 0.9059 & 0.7651          & 0.9399 & 0.3169          \\
                           & R-Adapter & 0.9261 & 0.8803          & 0.9296 & 0.2915          \\
                           & Ours      & 0.9151 & \textbf{0.4245}          & 0.9143 & \textbf{0.1241}   \\
                           \bottomrule
\end{tabular}
}
\end{table}

\subsection{Style Backdoor Attack}
Style backdoor attack \cite{qi-etal-2021-mind} is a hard attack challenge that our method is not able to solve perfectly. We evaluate our methods and previous baseline methods on three datasets under the mixed-label ($p=0.1$) and clean-label attack setups ($p=0.2$). The results are shown in Table \ref{tab:style_empirical}. We can find that most previous baseline methods cannot effectively defend against the style backdoor attack. Meanwhile, our method can only provide effective defense performance on AG's News dataset while the defense efficacy on the HSOL and SST-2 datasets is limited. The reason is that the words which could serve as a trigger is more complex given that the text style is the trigger. As a result, each base model of \name could still be misled by the backdoored sub-texts.

\begin{figure}[t]
  \centering
  \includegraphics[width=0.6\linewidth]{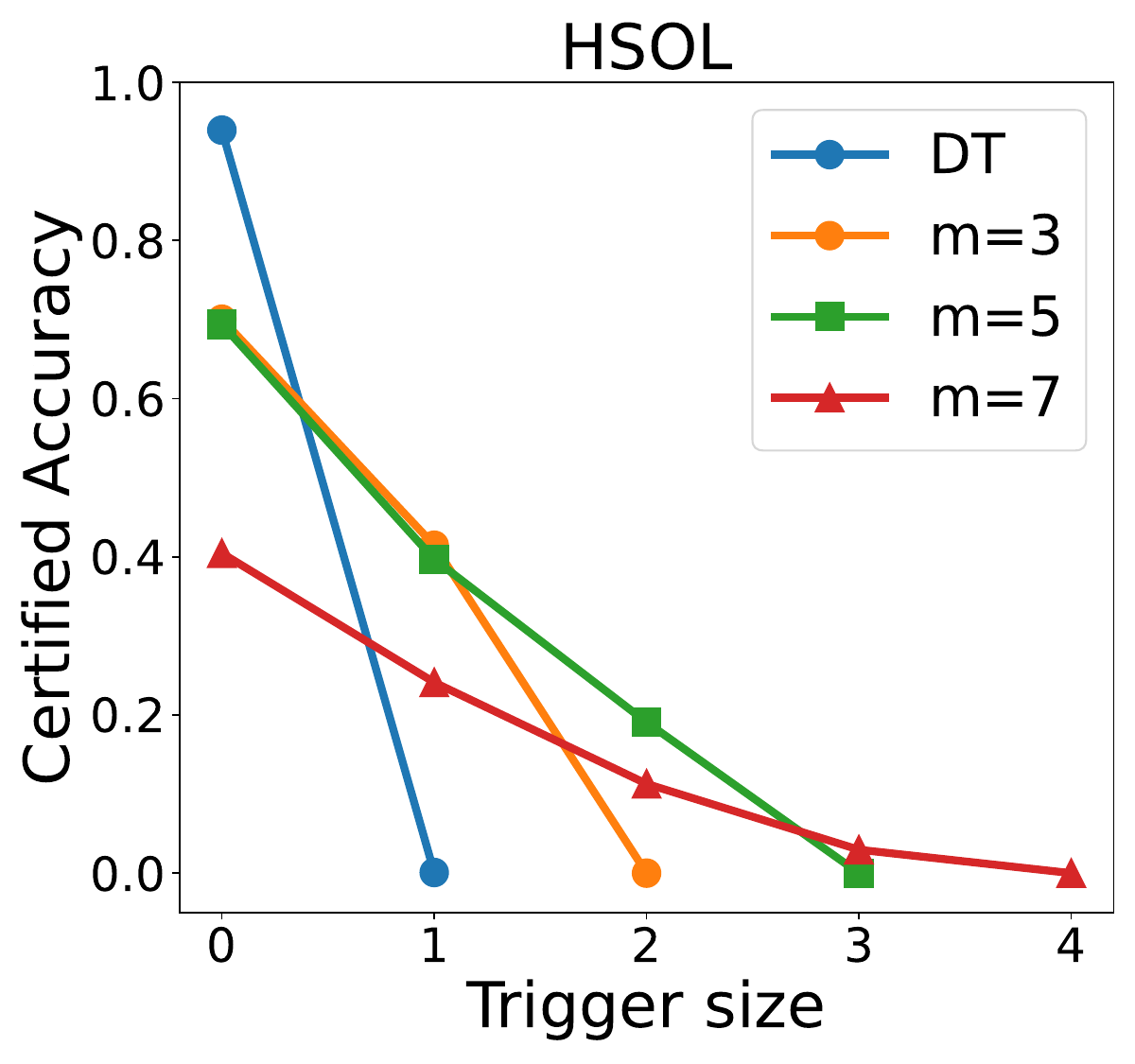}
  \caption{Certified results of \name under the dirty-label setups with $p=0.1$.}
  \label{fig:dirty_certified}
\end{figure}

\begin{table}[t]
\centering
\caption{Empirical performance of selected methods against dirty-label attacks with the poisoning rate $p=0.1$. }
\label{tab:dirty_empirical}
\resizebox{\linewidth}{!}{
\begin{tabular}{clcccccc}
\toprule
\multirow{2}{*}{Data} & \multicolumn{1}{c}{\multirow{2}{*}{Method}} & \multicolumn{2}{c}{BadNets}                        & \multicolumn{2}{c}{AddSent}                        & \multicolumn{2}{c}{SynBkd}                         \\
                      & \multicolumn{1}{c}{}                        & \multicolumn{1}{l}{CACC} & \multicolumn{1}{l}{ASR} & \multicolumn{1}{l}{CACC} & \multicolumn{1}{l}{ASR} & \multicolumn{1}{l}{CACC} & \multicolumn{1}{l}{ASR} \\
                      \midrule
\multirow{7}{*}{HSOL} & DT                                          & 0.9501                   & 0.9903                  & 0.9505                   & 0.9960                  & 0.9449                   & 0.9903                  \\
                      & ONION                                       & 0.9143                   & 0.7890                  & 0.9545                   & 1.0000                  & 0.9437                   & 0.9871                  \\
                      & BKI                                         & 0.9533                   & \textbf{0.0548}         & 0.9557                   & 1.0000                  & 0.9513                   & 0.9911                  \\
                      & STRIP                                       & 0.9569                   & 0.9992                  & 0.9581                   & 1.0000                  & 0.9919                   & 0.9919                  \\
                      & RAP                                         & 0.9549                   & 0.9992                  & 0.9573                   & 1.0000                  & 0.9529                   & 0.9823                  \\
                      & R-Adapter                                   & 0.8761 & 0.1739 & 0.8765 & 0.8068 & 0.8916 & 0.7576                \\
                      & Ours                                        & 0.8962                   & 0.1812                  & 0.9123                   & \textbf{0.1167}         & 0.9010                   & \textbf{0.5475}        \\
                      \bottomrule
\end{tabular}
}
\end{table}

\subsection{Dirty-Label Attack}
\label{app:dirty}
\update{We further evaluate \name against the \emph{dirty-label attack} where a backdoor attacker only poisons samples originally from the non-target class. 
We set the poisoning rate as $p=0.1$ and conduct certified evaluation and empirical evaluation on the HSOL dataset. 
The rest parameter setting is the same as Section~\ref{subsec:certified_setup} and Section~\ref{subsec:empirical_setup}.
}

\update{Figure~\ref{fig:dirty_certified} shows the certified results of \name on HSOL dataset under the dirty-label setup. 
Table~\ref{tab:dirty_empirical} shows the results of \name and the comparison baselines against word-level (BadWord, AddSent) and structure-level (SynBkd) attacks under the dirty-label attack setup. 
The findings are consistent with those in Section~\ref{subsec:certified_result} and Section~\ref{subsec:empirical_result}. }

\end{document}